# Factor-MCLS: Multi-agent learning system with reward factor matrix and multi-critic framework for dynamic portfolio optimization

Ruoyu Sun[a], Angelos Stefanidis[b], Zhengyong Jiang[c,*], Jionglong Su[d,*]

[a] School of AI and Advanced Computing, XJTLU Entrepreneur College (Taicang), Xi'an Jiaotong-Liverpool University, Suzhou, 215123, China, Ruoyu.Sun19@student.xjtlu.edu.cn, https://orcid.org/0009-0002-6052-0051

[b] School of AI and Advanced Computing, XJTLU Entrepreneur College (Taicang), Xi'an Jiaotong-Liverpool University, Suzhou, 215123, China, Angelos.Stefanidis@xjtlu.edu.cn, https://orcid.org/0000-0002-4703-8765

[c] School of AI and Advanced Computing, XJTLU Entrepreneur College (Taicang), Xi'an Jiaotong-Liverpool University, Suzhou, 215123, China, Zhengyong.Jiang02@xjtlu.edu.cn, https://orcid.org/0000-0001-8873-4073

[d] School of AI and Advanced Computing, XJTLU Entrepreneur College (Taicang), Xi'an Jiaotong-Liverpool University, Suzhou, 215123, China, Jionglong.Su@xjtlu.edu.cn, https://orcid.org/0000-0001-5360-6493

* Corresponding author.

# Abstract

Typical deep reinforcement learning (DRL) agents for dynamic portfolio optimization learn the factors influencing portfolio return and risk by analyzing the output values of the reward function while adjusting portfolio weights within the training environment. However, it faces a major limitation where it is difficult for investors to intervene in the training based on different levels of risk aversion towards each portfolio asset. This difficulty arises from another limitation: existing DRL agents may not develop a thorough understanding of the factors responsible for the portfolio return and risk by only learning from the output of the reward function. As a result, the strategy for determining the target portfolio weights is entirely dependent on the DRL agents themselves. To address these limitations, we propose a reward factor matrix for elucidating the return and risk of each asset in the portfolio. Additionally, we propose a novel learning system named Factor-MCLS using a multi-critic framework that facilitates learning of the reward factor matrix. In this way, our DRL-based learning system can effectively learn the factors influencing portfolio return and risk. Moreover, based on the critic networks within the multi-critic framework, we develop a risk constraint term in the training objective function of the policy function. This risk constraint term allows investors to intervene in the training of the DRL agent according to their individual levels of risk aversion towards the portfolio assets. We conduct experiments utilizing a portfolio comprising 29 stocks from the Dow Jones Index. In the training process, we demonstrate that our learning system can significantly enhance profitability while ensuring excellent risk control abilities. In the back-testing experiments, we refer to 14 strategies based on traditional capital growth theories and 10 strategies utilizing machine learning algorithms as comparative benchmarks. The results of the back-testing experiments demonstrate that our learning system outperforms the comparative strategies by at least 35.3% in profitability. Additionally, our learning system achieves at least 63.9% more in return per unit of risk when compared to other strategies.

*Keywords:* deep reinforcement learning, dynamic portfolio optimization, multi-critic, learning system

# 1. Introduction

In dynamic portfolio optimization, investors dynamically distribute funds among various assets in a portfolio according to their preferences for anticipated returns and associated risks. The goal is to achieve the highest possible returns while managing risk throughout each trading period by diversifying investments (Markowitz, 1952). Traditional portfolio optimization strategies can be classified into two primary categories: those based on Markowitz's mean-variance framework (Markowitz, 1952; Markowitz, 1959; Markowitz et al., 2000) and those grounded in Capital Growth Theory (Kelly, 1956; Hakansson & Ziemba, 1995). The strategies grounded in the Capital Growth Theory can be further subdivided into five distinct types, which include "Benchmarks," "Follow the Winner," "Follow the Loser," "Pattern-Matching Approaches," and "Meta-Learning Algorithms" (Li & Hoi, 2014).

The advancement of artificial intelligence (AI) technology has led to the adoption of various advanced AI techniques in dynamic portfolio optimization (Hambly et al., 2021), e.g., deep learning (DL), deep reinforcement learning (DRL), natural language processing (NLP), and generative adversarial networks (GANs). Among these techniques, DRL achieves significant results in exploring policies for dynamic portfolio optimization (Gao et al., 2023; Hambly et al., 2021; Gao et al., 2020; Gao et al., 2021; Sun et al., 2021; Zhang et al. 2024; Zhang et al., 2022; Gao et al., 2022; Song et al., 2023; Ren et al., 2021; Gu et al., 2021). In the DRL algorithm, the dynamic portfolio optimization in consecutive trading periods is assumed to be a Markov decision process (MDP). As a model-free approach, the DRL agent does not need specialized learning of the price dynamics of assets within a portfolio. Instead, it focuses on exploring dynamic portfolio strategies that maximize the reward function within the training environment. In recent years, many attempts have been made to apply different DRL-based methods to explore the policy in dynamic portfolio optimization. Jiang et al. (2017) proposed an EIIE framework for constructing the policy network for portfolio optimization. Zhang et al. (2020) introduced a DRL-based cost-sensitive portfolio selection model, which comprises a cost-sensitive Portfolio Policy Network (PPN) that employed a two-stream network framework to integrate price data and insights on asset correlations. FinRL (Liu et al., 2021) constructed an open-source library that includes the application of several state-of-the-art DRL algorithms for making trading decisions in portfolio optimization, which included deep deterministic policy gradient (DDPG) (Lillicrap et al., 2015), soft actor-critic (SAC) (Haarnoja et al., 2018), Twin Delayed Deep Deterministic policy gradient (TD3) (Fujimoto et al., 2018), Proximal Policy Optimization (PPO) (Schulman et al., 2017)), and so on. Stockformer (Gao et al., 2023) incorporated a predictive module into the actor-critic reinforcement learning (RL) framework. This approach combines the strengths of existing stock models for learning future dynamics with the advantages of DRL-for-finance techniques in developing more adaptable policies. TC-Mac (Yang, 2023) effectively identified the relationships between the characteristics of local assets and the embeddings of the global context through the optimization of mutual information. BDA (Sun et al., 2024) integrated the traditional Black-Litterman model with a DRL agent to develop a long/short trading strategy for portfolio optimization in consecutive trading periods. Although DRL algorithms have been widely used in dynamic portfolio optimization, inspired by the idea of Sun et al. (2025), we find that, when applying the actor-critic algorithm and deep function approximators to train the DRL agent for dynamic portfolio optimization, it is difficult to significantly improve the profitability and risk-adjusted profitability of our DRL agent in the training process. Consequently, researchers are unable to demonstrate that the policy function used in the out-of-sample environment is learned from a thorough exploration of optimal strategies within the training environment. To overcome this issue, Sun et al. (2025) propose a multi-agent learning system where an auxiliary agent is adopted to assist the executive agent in determining the target portfolio weights. Despite partially addresses this issue, the DRL agents designed for dynamic portfolio optimization still have two major research gaps:

1) When the construction of the reward function is based on the risk-adjusted return, the reward function calculation involves numerous influencing factors, including the price changes of each portfolio asset, the variance of each portfolio asset's price fluctuations, and the correlation coefficients between price fluctuations of different assets. It is challenging for a DRL agent to learn all the factors affecting the reward value by learning the output values of the reward function.

2) For the DRL agents currently used in dynamic portfolio optimization, investors cannot intervene in the agents' training based on their risk preferences for individual assets within the portfolio. This lack of intervention makes aligning the training outcomes with investor objectives difficult.

To address these research gaps, we propose using a novel reward factor matrix in the training environment to identify the factors in the reward function. Furthermore, we construct a novel DRL-based learning system with a multi-critic framework to learn the factors in the reward factor matrix. Here, the learning system is referred to as the reward factor learning-based multi-critics learning system (Factor-MCLS). By applying the reward factor metric and multi-critic framework, the DRL agent in our learning system can thoroughly learn the factors in the reward function. Hence, the first research gap proposed above can be overcome. In the training objective function of the policy function in our learning system, we construct a risk constraint term based on the critic networks in the multi-critic framework to control each asset's position size based on the investor's risk aversion to each portfolio asset. Hence, investors can intervene in the training of the DRL agent based on their subjective risk aversion of individual portfolio assets. Moreover, to ensure the efficiency of our training algorithm, we borrow the idea of Sun et al. (2025) to construct a module to track the indices' trajectories. The trajectories reflect the DRL agent's profitability and risk control ability in the training environment (i.e. the return value, variance, and risk-adjusted return).

The novelties of our work are threefold:
1) We propose using a reward factor matrix to describe the influencing factors in the reward function calculation process. Furthermore, we construct a multi-critic framework in our learning system (Factor-MCLS) to learn all factors in the reward factor matrix. Hence, DRL agents in Factor-MCLS can achieve a thorough understanding of the factors influencing the reward function value in the training process.
2) Based on the multi-critic framework, we construct a risk constraint term in the training objective function of the policy network to reduce the computational cost the agent spends on exploring action spaces with negative returns and high variance. This enhances the agent's efficiency in exploring the environment.
3) Within the framework of the risk constraint term, investors can intervene in the training of the DRL agent based on their subjective risk aversion of individual stocks by determining the risk aversion vector in the risk constraint term. This ensures that the training objective of the DRL agent aligns closely with the investor's personal risk aversion.

The key contributions of this paper are as follows:
1) In the training process, we combine our novel multi-critic framework with DDPG (Lillicrap et al., 2015) algorithm to train the critics within the multi-critic framework and the policy network in the learning system. The trajectories of the training objective functions value for the critics and actor networks indicate effective convergence in the training process. It illustrates that such a network update method can effectively avoid gradient vanishing and explosion issues.
2) We construct a novel multi-agent learning system (Factor-MCLS) which can comprehensively learn the factors defined in the reward factor matrix. In out-of-sample experiments, our Factor-MCLS achieves a significant advantage in profitability per unit of risk compared to benchmark strategies. This suggests that the policies learned by Factor-MCLS in the training environment can maintain outstanding generalization abilities in out-of-sample scenarios.

3) During the training process, we track the indices within the training environment, which reflect Factor-MCLS's profit and risk control abilities. By tracking these indices, we observe that its profit and risk control abilities are significantly improved in the training process. Hence, we conclude that the critic networks within the proposed multi-critic framework can learn the factors within the reward factor matrix and train Factor-MCLS to obtain more rewards from the environment by providing gradient propagation to the policy network. Hence, Factor-MCLS's profitability and risk control ability are effectively enhanced.

The remainder of this paper is organized as follows. Section 2 outlines the preliminary work conducted in support of our research. Section 3 defines the MDP within the dynamic portfolio optimization. Section 4 details the methodology employed in our study. Section 5 presents the empirical findings derived from our experiments. Finally, the conclusions and future research directions are given in Section 6.

## 2. Preliminaries
### 2.1 Basic concepts

**Trading period.** A trading period is the minimum time unit for reallocating funds among assets in a portfolio. In this context, the reallocation of funds takes place at the conclusion of the selected trading day. Research by Li et al. (2023) suggests that high-frequency trading data typically displays a diminished signal-to-noise ratio. To prevent the DRL agent from learning too much noisy information from the environment, we extend the trading period, thereby reducing the trading frequency. Consequently, commencing with the initial fund allocation action, fund reallocation actions are implemented every $K$ trading days over the investment horizon.

As shown in Figure 1, a trading period commences with the action allocation at the end of a trading day. It concludes before the next action reallocation, which occurs after $K$ trading days. Here, the $t^{\text{th}}$ trading period is defined as the time interval $[t, t+1), t = 1, 2, \ldots, T_f$, where $T_f$ is the number of trading periods in the entire investment horizon. During a trading period, there are a total of $K$ trading days. Here, the specific trading day $t_k$ corresponds to the $k^{\text{th}}$ day of the $t^{\text{th}}$ trading period.

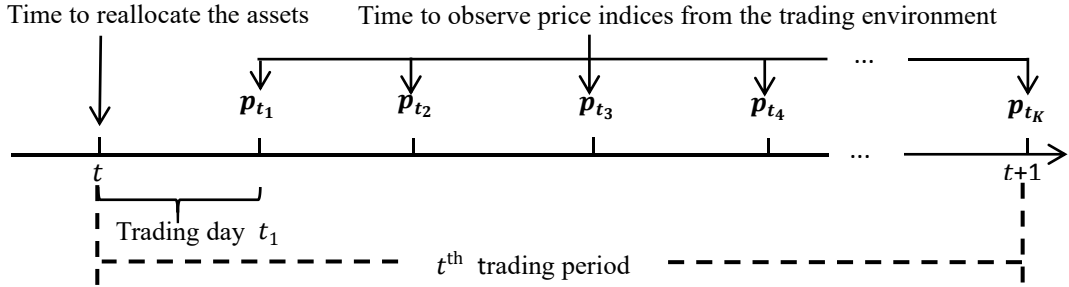

**Fig 1:** The designated observation intervals within the $t^{\text{th}}$ trading period. In the $t^{\text{th}}$ trading period, there are $K$ distinct trading days $t_k, k = 1, 2, \ldots K$. At the end of each trading day $t_k, k = 1, 2, \ldots, K$, the price vector $\boldsymbol{p}_{t_k}, k = 1, 2, \ldots, K$ is observed for the description of the state in the Markov Decision Process (MDP).

**Price vector $\boldsymbol{p}_{t_k}$.** To describe the price fluctuations of individual portfolio assets, we collect the adjusted closing prices of each portfolio asset at the end of each trading day to construct the price vector $\boldsymbol{p}_{t_k} \in \mathbb{R}^{n \times 1}$, defined as
$$\boldsymbol{p}_{t_k} = [p_{1,t_k}, p_{2,t_k}, \ldots, p_{n,t_k}]^T, \tag{1}$$
where $p_{i,t_k}$ is the adjusted closing price of the asset $i$ on the $k^{\text{th}}$ trading day in the $t^{\text{th}}$ trading period, and $n$ is the number of assets in the portfolio. Furthermore, the price vector on the last trading day $K$ in the $t^{\text{th}}$ trading period is also defined as $\boldsymbol{p}_t$:
$$\boldsymbol{p}_t = \boldsymbol{p}_{t_K} = [p_{1,t}, p_{2,t}, \ldots, p_{n,t}]^T.$$
Here, adjusted closing price is used to describe asset price changes because it accounts for factors such as dividends, stock splits, and other corporate actions that can affect the closing price of a security (Norton, 2011). By adjusting for these factors, the adjusted closing price provides a more accurate representation of the true price movements of the asset over time. Hence, in our research, the value of each asset in the portfolio is measured using the adjusted closing price.

**Relative price ratio vector $\boldsymbol{z}_{t_k}$.** To accurately describe changes in asset prices, we construct the relative price ratio vector $\boldsymbol{z}_{t_k} \in \mathbb{R}^{n \times 1}$ as follows:
$$\boldsymbol{z}_{t_k} = \begin{cases} \boldsymbol{p}_{t_k} \oslash \boldsymbol{p}_{t_{k-1}} & k = 2, 3, 4, \ldots, K \\ \boldsymbol{p}_{t_k} \oslash \boldsymbol{p}_{t-1} & k = 1 \end{cases}, \tag{2}$$
where $\oslash$ represents the element-wise division.

**Price fluctuation matrix $\boldsymbol{U}_t$.** To provide a more comprehensive description of the state for the DRL agents, we construct a price fluctuation tensor $\boldsymbol{U}_t \in \mathbb{R}^{K \times n}$ that represents the price fluctuation over the entire trading period. Price fluctuation tensor. $\boldsymbol{M}_t$ is a matrix which comprises all the relative price ratio vectors within the $t^{\text{th}}$ trading period. The price fluctuation matrix in the $t^{\text{th}}$ trading period is denoted as
$$\boldsymbol{U}_t = [\boldsymbol{z}_{t_1}, \boldsymbol{z}_{t_2}, \ldots, \boldsymbol{z}_{t_K}],$$
where $\boldsymbol{z}_{t_k}$ is the relative price ratio defined in Equation (2).

**Long position and short position.** Here, we assume that investors can simultaneously implement long and short strategies in the current market. Specifically, a long position refers to the ownership of an asset with the expectation that its value will increase over time. This means that the investor or trader buys the asset intending to sell it at a higher price in the future to make a profit. A short position involves selling an asset the trader does not own in the hope that its price will decrease. The trader then aims to buy back the asset at a lower price in the future to cover the initial sale, thus profiting from the price difference.

**Target portfolio weights $\boldsymbol{w}_t$.** In dynamic portfolio optimization, the primary aim is to determine the target portfolio weights vector $\boldsymbol{w}_t \in \mathbb{R}^{1 \times n}$ in each trading period, defined as:

$$\boldsymbol{w_t} = [w_{1,t}, w_{2,t}, \ldots, w_{n,t}]^T,$$

where $w_{i,t}$ is the target weight of the asset $i$ in the $t^{\text{th}}$ trading period.

In real-world trading scenarios, the target position sizing of each portfolio asset $\boldsymbol{q_t}$ is calculated based on the target portfolio weights $\boldsymbol{w_t}$ and investment amount $T_t$:

$$\boldsymbol{q_t} = [q_{1,t}, q_{2,t}, \ldots, q_{n,t}]^T = \lfloor (T_t \boldsymbol{w_t}) \oslash \boldsymbol{p_{t-1}} \rfloor, \tag{3}$$

where $\boldsymbol{p_{t-1}}$ is the closing price vector at the end of the $t-1^{\text{th}}$ trading period, and $\oslash$ denotes the element-wise division. Given the assumption that a single share of stock cannot be divided, we employ the floor function $\lfloor \cdot \rfloor$ in Equation (3) to ensure that the elements within the vector of the target position $\boldsymbol{q_t}$ are integers. To attain the target position, it is also necessary to compute the market order $\boldsymbol{\Delta q_t}$, which represents the difference in target quantities between two adjacent trading periods:

$$\boldsymbol{\Delta q_t} = [\Delta q_{1,t}, \Delta q_{2,t}, \ldots, \Delta q_{n,t}]^T = \boldsymbol{q_t} - \boldsymbol{q_{t-1}}.$$

**Portfolio value $v_{t_k}^p$ and total assets value $v_{t_k}^{total}$.** As we assume that the investment amount at each trading period is $T_t$, the portfolio value and total assets value are not equal throughout the entire investment horizon. Specifically, the total assets value $v_{t_k}^{total}$ is the sum of all risky assets and cash. Hence, the value of total assets at the end of the $k^{\text{th}}$ trading day within the $t^{\text{th}}$ trading period is calculated as:

$$v_{t_k}^{total} = c_t + \boldsymbol{q_t}^T \boldsymbol{p_{t_k}}, \tag{4}$$

where $c_t$ is the value of cash in the $t^{\text{th}}$ trading period, $\boldsymbol{q_t}$ and $\boldsymbol{p_{t_k}}$ are, respectively, as defined in Equations (3) and (1).

For the portfolio value $v_{t_k}^p$, at the beginning of each trading period, the portfolio value equals the investment amount. At the end of each trading day within the trading period, the portfolio value $v_{t_k}^p$ can be calculated based on the value changes in the total assets:

$$v_{t_k}^p = T_t + (v_{t_k}^{total} - v_{(t-1)_K}^{total}),$$

where $v_{t_k}^{total}$ is as defined in Equation (4), and $T_t$ is as defined in Equation (3).

*2.2 Assumptions*

This research exclusively employs back-testing methodology to execute all experiments, wherein investors initiate trading from a particular historical time period without any foreknowledge of future market circumstances. To adhere to the criteria of the back-tested trading scenarios in each experiment, we define the assumptions (A1) and (A2). Furthermore, Since the policy learned by our constructed learning system is based on long-short operations, we additionally propose the Assumption (A3). The specific assumptions are given as follows:

(A1) Zero Slippage: It is assumed that the trading decisions made at the start of each trading period have minimal influence on the movement of asset prices.
(A2) Zero Market Impact: The assets within the portfolio are selected based on their high market liquidity, allowing for prompt execution of all trading orders immediately upon issuance.
(A3) Short Selling Permit: No restrictions are placed on the practice of short selling within the specified market.

In a realistic trading environment, the aforementioned assumptions (A1) and (A2) remain valid when the trading volume of the chosen assets in the portfolio is significantly large. To satisfy assumption (A3), we conduct experiments within the equity market, permitting long and short-selling trades (e.g., the United States stock market).

# 3. Dynamic portfolio optimization in consecutive trading periods as MDPs

The dynamic portfolio optimization problem in consecutive trading periods is treated as an MDP with a 5-element tuple:

$$\text{MDP} = <\mathcal{S}, \mathcal{A}, \mathcal{P}, \mathcal{R}, \gamma, \mathcal{T}>,$$

where $\mathcal{S}$ denotes the state space, $\mathcal{A}$ denotes the action space, $\mathcal{P}: \mathcal{S} \times \mathcal{A} \to \mathcal{S}$ denotes the state transition dynamics. Specifically, it is defined as a probability distribution $P(\boldsymbol{s_{t+1}}|\boldsymbol{s_t}, \boldsymbol{a_t})$. $\mathcal{R}$ is the reward function, and $\gamma$ is the discount factor. Although the decision process of the portfolio optimization problem in consecutive trading periods is infinite in the real-world scenario, here, we define the problem of dynamic portfolio optimization within a fixed time period as our target MDP. Hence, the length of the trading periods in the environment is set as a fixed value $T_f^{tr}$. $\mathcal{T}$ is the set of terminal states. The detailed definition of the action space $\mathcal{A}$, state space $\mathcal{S}$, and reward function $\mathcal{R}$ are given as follows:

*Action space $\mathcal{A}$.* Here, the DRL agent for portfolio optimization has a continuous action space, such that $\boldsymbol{a_t} \in \mathbb{R}^{n \times 1}$, where the action $\boldsymbol{a_t}$ at the beginning of the $t^{\text{th}}$ trading period is defined as the target portfolio weight $\boldsymbol{w_t}$ of each asset in the portfolio:

$$\mathcal{A}: \boldsymbol{a_t} = \boldsymbol{w_t}.$$

*State space $\mathcal{S}$.* Following the idea of Hernandez et al. (2021), we adopt an auxiliary agent to accomplish an auxiliary task that determines the baseline portfolio weights to assist our DRL agent in determining the target portfolio weights $\boldsymbol{w_t}$. Here, we apply the pre-trained BL model based deep reinforcement learning agent proposed by Sun et al. (2025) to undertake this auxiliary task. Hence, the state of our DRL agent is described by two kinds of data, i.e., the data collected from the market and baseline portfolio weights determined by the auxiliary agent.

Here, the market data is described by the historical price ratio matrix $\boldsymbol{X_t} \in \mathbb{R}^{n \times KM}$ constructed by the relative price ratio matrix $\boldsymbol{U_t}$ in the past $M$ trading periods:

$$\boldsymbol{X_t} = [\boldsymbol{U_{t-M}}, \ldots, \boldsymbol{U_{t-2}}, \boldsymbol{U_{t-1}}].$$

By combining the price ratio matrix $\boldsymbol{X_t}$ and the baseline portfolio weights $\boldsymbol{w_t^{au}}$ determined by the auxiliary agent, the state $\boldsymbol{s_t}$ is described by a two-element tuple:

$$\boldsymbol{s_t} = \langle \boldsymbol{X_t}, \boldsymbol{w_t^{au}} \rangle.$$

*Reward function $\mathcal{R}$.* The reward function $\mathcal{R}: r(\boldsymbol{a_t}, \boldsymbol{s_t}|\lambda_1, \lambda_2)$ is defined as the average daily return of the portfolio modified by the corresponding variance and transaction scale ratio:

$$\mathcal{R}: r(\boldsymbol{a_t}, \boldsymbol{s_t}|\lambda_1, \lambda_2) = \frac{1}{K}\left[\frac{v_{t_K}^p}{T_t} - 1\right] - \lambda_1 [\boldsymbol{w_t}^T \boldsymbol{\Sigma_t} \boldsymbol{w_t}] - \lambda_2 \left[\frac{|\boldsymbol{\Delta q_t}|^T \boldsymbol{p_{t-1}}}{T_t}\right], \tag{5}$$

where $v_{t_K}^p$ is the value of the portfolio at the end of the last trading day within the $t^{\text{th}}$ trading period just before the asset reallocation. In Equation (5), the term $\frac{1}{K}\left[\frac{v_{t_K}^p}{T_t} - 1\right]$ denotes the relative daily return of the portfolio in the $t^{\text{th}}$ trading period. The term $\boldsymbol{w_t}^T \boldsymbol{\Sigma_t} \boldsymbol{w_t}$ is the calculation of the variance of the portfolio, where the covariance matrix $\boldsymbol{\Sigma_t} \in \mathbb{R}^{n \times n}$ is calculated based on the relative return of the portfolio assets in the $t^{\text{th}}$ trading period and last $m$ trading period:

$$\Sigma_t = \frac{\left[X_t - 1 - \frac{1}{KM}\sum_{i=1}^{M}\sum_{k=1}^{K}(z_{t-i_k} - 1')\right]\left[X_t - 1 - \frac{1}{KM}\sum_{i=1}^{M}\sum_{k=1}^{K}(z_{t-i_k} - 1')\right]^T}{KM - n - 1},$$

where $1 \in \mathbb{R}^{n \times KM}$ is a matrix of all ones and $1' \in \mathbb{R}^{n \times 1}$ is a vector of all ones. In Equation (5), the term $\frac{|\Delta q_t|^T p_{t-1}}{T_t}$ describes the transaction scale caused by the portfolio weights $w_t$, where $\Delta q_t$, $p_{t-1}$, and $T_t$ are as given in Equation (3). Since we need to control the variance of the portfolio and the transaction scale, in Equation (5), the parameters $\lambda_1$ and $\lambda_2$ should be larger than zero.

## 4. Methodology

### 4.1 Training logic of the DRL agents in our Factor-MCLS

Borrowing the idea of Sun et al. (2025), an auxiliary agent (i.e., BDA proposed by Sun et al. (2024)) and our DRL agent constitute a multi-agent learning system (i.e., Factor-MCLS) to explore the optimal policy for portfolio optimization. Furthermore, we employ a hierarchical deep reinforcement learning (HDRL) algorithm (Barto & Mahadevan, 2003) to train the agents in the learning system. By applying hierarchical reinforcement learning, the portfolio optimization problem is decomposed into a series of sub-problems. Compared to the original problem before hierarchical decomposition, these sub-problems exhibit reduced complexity. In the training process, we first train the auxiliary agent based on the deep policy gradient algorithm proposed by Sun et al.(2024). After completing the training of the auxiliary agent, our DRL agent learns the optimal strategy in the environment with the assistance of the auxiliary agent. In its decision-making process, the auxiliary agent first outputs baseline portfolio weights $w_t^{au}$ based on market data. Subsequently, our DRL agent outputs target portfolio weights based on the baseline portfolio weights and market data. The specific decision logic of our learning system Factor-MCLS is given in Figure 2.

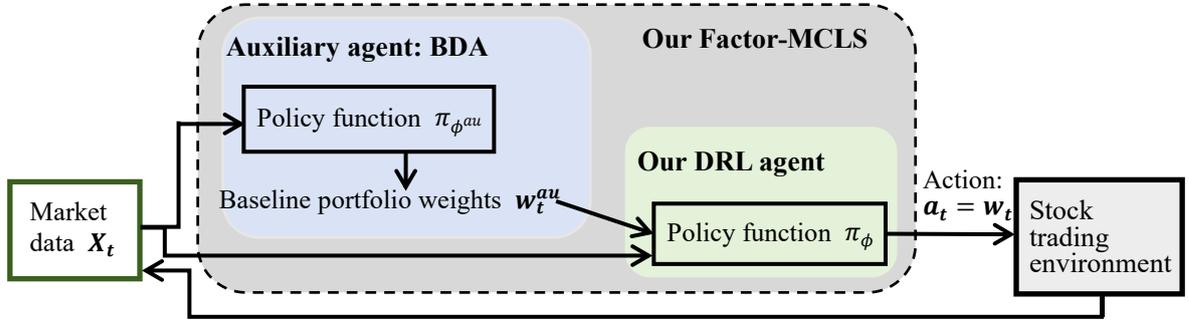

**Fig 2:** The illustration of the decision logic of the target portfolio weights in our Factor-MCLS. Except for receiving the state $s_t$ from the environment, the auxiliary agent BDA also provides baseline portfolio weights $w_t^{au}$ to assist our DRL agent in determining the target portfolio weights $w_t$. In the decision logic of the target portfolio weights $w_t$, the auxiliary agent BDA outputs the baseline portfolio weights $w_t^{au}$ based on its policy function $\pi_{\phi^{au}}$. Based on the baseline portfolio weights $w_t^{au}$ and the market data $X_t$, our DRL agent determines the target portfolio weights $w_t$ and takes action (i.e., makes market orders) to achieve the target portfolio weights $w_t$.

### 4.2 Reward factor matrix

The reward function $r(a_t, s_t | \lambda_1, \lambda_2)$ in the environment is as defined in Equation (5). In calculating this reward function value, we simultaneously consider the return, risk (including variance and covariance), and transaction scale of each asset in the portfolio. As a result, the calculation of the reward function is relatively complex, which poses challenges for the critic network in learning state-action values. Here, we enable our DRL agent to consider all factors that influence the value of the reward function in the training process based on the actor-critic algorithm. Therefore, our Factor-MCLS can develop a corresponding dynamic portfolio optimization policy based on the investor's subjective risk aversion for various portfolio assets. Apart from computing the reward based on the reward function defined in Equation (5), we also construct a reward factor matrix $r_{elem}$ in the environment to characterize the impact of each portfolio asset on the profit, risk, and transaction scale contributed to the portfolio. Specifically, the reward factor matrix $r_{elem}$ consists of four different factor vectors, i.e., the return factor vector $r_t^{Re}$, variance factor vector $r_t^{Va}$, covariance factor vector $r_t^{Co}$, and transaction scale factor vector $r_t^{Ts}$:

$$r_t^{elem}: r_{elem}(a_t, s_t | \lambda_1, \lambda_2) = \left[ r_t^{Re}, r_t^{Va}, r_t^{Co}, r_t^{Ts} \right].$$

Hence, different from the traditional actor-critic algorithm, the tuple stored in the reply buffer in the training process can be defined as:

$$(s_t, a_t, r_t^{elem}, r_t, s_{t+1}).$$

To facilitate the learning process of the DRL agent and to prevent challenges arising from excessively small data values in the variance-covariance matrix, it is essential to utilize percentage-based return data $\varpi_t \in \mathbb{R}^{n \times KM}$ when calculating the elements in the reward factor matrix $r_t^{elem}$:

$$\varpi_t = [[U_{t-M}, \ldots, U_{t-2}, U_{t-1}] - 1] \cdot 100\%,$$

where $1 \in \mathbb{R}^{n \times KM}$ is a matrix of all ones. This approach applies specifically to the computations of the return factor vector $r_t^{Re}$, variance factor vector $r_t^{Va}$, and covariance factor vector $r_t^{Co}$. Furthermore, the variance-covariance matrix derived from the percentage-based return vector $\varpi_t$ is defined as follows:

$$\varpi(\Sigma_t) = \frac{\left[\varpi_t - \frac{1}{KM}\sum_{i=1}^{M}\sum_{k=1}^{K}(z_{t-i_k} - 1') \cdot 100\%\right]\left[\varpi_t - \frac{1}{KM}\sum_{i=1}^{M}\sum_{k=1}^{K}(z_{t-i_k} - 1') \cdot 100\%\right]^T}{KM - n - 1} = \begin{bmatrix} \sigma'^2_{t,11} & \sigma'_{t,12} & \cdots & \sigma'_{t,1n} \\ \sigma'_{t,21} & \sigma'^2_{t,22} & \cdots & \sigma'_{t,2n} \\ \vdots & & \ddots & \vdots \\ \sigma'_{t,n1} & \sigma'_{t,n2} & \cdots & \sigma'^2_{t,nn} \end{bmatrix},$$

where the element $\sigma'^2_{t,ii}$ represents the variance of the asset $i$ in the $t^{th}$ trading period, while the element $\sigma'_{t,ij}$ denotes the covariance between asset $i$ and asset $j$ in the $t^{th}$ trading period. The symbol $\mathbf{1'} \in \mathbb{R}^{n \times 1}$ denotes the vector of all ones. Here, we give the detailed definition of each factor vector in the reward factor matrix as follows:

- In the reward factor matrix $r_t^{elem}$, the return factor vector $r_t^{Re} \in \mathbb{R}^{n \times 1}$ is constructed to represent the return contribution of each stock in the portfolio:

$$r_t^{Re} = [r_{t,1}^{Re}, r_{t,2}^{Re}, \ldots, r_{t,n}^{Re}]^T$$
$$= r_{Re}(a_t, s_t) = \frac{1}{K}\left[\frac{q_t \otimes (p_t - p_{t-1}) - \alpha |\Delta q_t| \otimes p_{t-1}}{T_t}\right] \cdot 100\%, \quad (6)$$

where the factor $r_{t,i}^{Re}$ represents the return contribution of the portfolio asset $i$ in the $t^{th}$ trading period, and $\otimes$ represents the element-wise multiple. Hence, the sum of total elements in the return factor vector is nearly equivalent to the value of the term $\frac{1}{K}\left[\frac{v_{t_K}^p}{T_t} - 1\right]$ in the reward function defined in Equation (5).

- Variance factor vector $r_t^{Va} \in \mathbb{R}^{n \times 1}$ in the reward factor matrix is adopted to represent the variance contribution of each stock in the portfolio when only considering the variance of each stock:

$$r_t^{Va} = [r_{t,1}^{Va}, r_{t,2}^{Va}, \ldots, r_{t,n}^{Va}]^T$$
$$= r_{Va}(a_t, s_t) = [w_{t,1}^2 \sigma'^2_{t,11}, w_{t,2}^2 \sigma'^2_{t,22}, \ldots, w_{t,n}^2 \sigma'^2_{t,nn}]^T, \quad (7)$$

where the element $r_{t,i}^{Va}$ denotes the risk hedged by the covariance between the portfolio asset $i$ and the other portfolio asset in the $t^{th}$ trading period.

- Covariance factor vector $r_t^{Co} \in \mathbb{R}^{n \times 1}$ is adopted to represent the volatility contribution of each stock based on the covariance:

$$r_t^{Co} = [r_{t,1}^{Co}, r_{t,2}^{Co}, \ldots, r_{t,n}^{Co}]^T$$
$$= r_{Co}(a_t, s_t) = w_t \otimes (\varpi(\Sigma_t)w_t) - [w_{t,1}^2 \sigma'^2_{t,11}, w_{t,2}^2 \sigma'^2_{t,22}, \ldots, w_{t,n}^2 \sigma'^2_{t,nn}]^T, \quad (8)$$

where the element $r_{t,i}^{Co}$ denotes the risk caused by the variance of the portfolio asset $i$ in the $t^{th}$ trading period. Hence, the sum of the elements in the variance factor vector $r_t^{Va}$ and covariance factor vector $r_t^{Co}$ is equal to the value of the term $w_t^T \Sigma_t w_t$ in the reward function defined in Equation (5).

- Transaction scale factor vector $r_t^{Ts} \in \mathbb{R}^{n \times 1}$ is adopted to represent the transaction scale contribution of each stock in the portfolio:

$$r_t^{Ts} = [r_{t,1}^{Ts}, r_{t,2}^{Ts}, \ldots, r_{t,n}^{Ts}]^T$$
$$= r_{Ts}(a_t, s_t) = \left[\frac{|\Delta q_t| \otimes p_{t-1}}{T_t}\right], \quad (9)$$

where the element $r_{t,i}^{Ts}$ represents the transaction scale caused by the portfolio asset $i$ in the $t^{th}$ trading period. Similar to the return factor vector, the sum of total elements in the transaction scale factor vector is equal to the value of the term $\frac{|\Delta q_t|^T p_{t-1}}{T_t}$ in the reward function defined in Equation (5).

*4.3 Multi-critic framework*

Here, we adopt the deep deterministic policy gradient (DDPG) (Lillicrap et al., 2015) algorithm to train our DRL agent to learn the optimal policy $\pi_{\theta^*}$. DDPG is among the most famous actor-critic algorithms for deterministic policy. Since its training objective is maximizing our DRL agent's action value, we can balance long-term and short-term returns in the training process. In traditional DDPG algorithm (Lillicrap et al., 2015), the output of the reward function in the environment is a single value. Since the critic network is adopted to learn the state-action value function (Lillicrap et al., 2015), the output of the critic network is also a single value. However, in our framework, we train our DRL agent to learn all the information in the reward factor matrix $r_t^{elem}$. Different from the traditional actor-critic algorithm, we define four action-value functions $Q_{Re}^{\pi_\theta}(s_t, a_t), Q_{Va}^{\pi_\theta}(s_t, a_t), Q_{Co}^{\pi_\theta}(s_t, a_t),$ and $Q_{Ts}^{\pi_\theta}(s_t, a_t)$, responding to four factor vectors $r_t^{Re}, r_t^{Va}, r_t^{Co}$, and $r_t^{Ts}$ in the reward factor matrix $r_{elem}$. Hence, the output of these four action-value functions is an $n \times 1$ vector, responding to each element in the factor vector. In our DRL algorithm, four critic networks are, respectively, adopted to learn the action-value functions of four factor vectors (i.e., $r_t^{Re}, r_t^{Va}, r_t^{Co}$, and $r_t^{Ts}$) in the reward factor matrix $r_t^{elem}$.

The detailed definition of each action-value function and the corresponding critic network in the multi-critic framework are given as follows:

- The critic network $Q_{\omega_{(Re)}}(s_t, a_t) \in \mathbb{R}^{n \times 1}$ with parameters $\omega_{(Re)}$ is adopted to approximate the action-value $Q_{Re}^{\pi_\theta}(s_t, a_t)$ of the elements in the return factor vector $r_t^{Re}$. Here, the action-value function $Q_{Re}^{\pi_\theta}(s_t, a_t)$ is defined as:

$$Q_{Re}^{\pi_\theta}(s_t, a_t) \triangleq [q_1^{Re}, q_2^{Re}, \ldots, q_n^{Re}]^T = \mathbb{E}_{s_{i>t}, r_{i \geq t}^{elem}, a_{i>t} \sim \pi_\theta}\left[\sum_{i=t}^{\mathcal{T}} \gamma^{i-t} r_t^{Re} | s_t, a_t\right], \quad (10)$$

where $r_t^{Re}$ is the return factor vector defined in Equation (6), and $\gamma$ is the discount rate defined in the Markov decision process, and $\mathbb{E}$ represents the expected value function. Here, the $i^{th}$ element in the output of the critic network $Q_{\omega_{(Re)}}(s_t, a_t)$ is defined as $Q_{\omega_{(Re)}}^i(s_t, a_t)$.

- Critic network $Q_{\omega_{(Va)}}(s_t, a_t) \in \mathbb{R}^{n \times 1}$ with parameters $\omega_{(Va)}$ is adopted to approximate the action-value $Q_{Va}^{\pi_\theta}(s_t, a_t)$ of the elements in the variance factor vector $r_t^{Va}$, which is defined as:

$$Q_{Va}^{\pi_\theta}(s_t, a_t) \triangleq [q_1^{Va}, q_2^{Va}, \ldots, q_n^{Va}]^T = \mathbb{E}_{s_{i>t}, r_{i \geq t}^{elem}, a_{i>t} \sim \pi_\theta}\left[\sum_{i=t}^{\mathcal{T}} \gamma^{i-t} r_t^{Va} | s_t, a_t\right],$$

where $r_t^{Va}$ is the variance factor vector defined in Equation (7), and $\gamma$ is as defined in Equation (10). Here, the $i^{th}$ element in the output of the critic network $Q_{\omega_{(Va)}}(s_t, a_t)$ is defined as $Q_{\omega_{(Va)}}^i(s_t, a_t)$.

- Critic network $Q_{\omega_{(Co)}}(s_t, a_t) \in \mathbb{R}^{n \times 1}$ with parameters $\omega_{(Co)}$ is adopted to approximate the action-value $Q_{Co}^{\pi_\theta}(s_t, a_t)$ of the elements in the covariance factor vector $r_t^{Co}$, which is defined as:

$$Q_{Co}^{\pi_\theta}(s_t, a_t) \triangleq [q_1^{Co}, q_2^{Co}, \ldots, q_n^{Co}]^T = \mathbb{E}_{s_{i>t}, r_{i \geq t}^{elem}, a_{i>t} \sim \pi_\theta}\left[\sum_{i=t}^{\mathcal{T}} \gamma^{i-t} r_t^{Co} | s_t, a_t\right],$$

where $r_t^{Co}$ is the covariance factor vector defined in Equation (8), and $\gamma$ is as defined in Equation (10). Here, the $i^{th}$ element in the output of the critic network $Q_{\omega_{(Co)}}(s_t, a_t)$ is defined as $Q_{\omega_{(Co)}}^i(s_t, a_t)$.

- Critic network $Q_{\omega_{(Ts)}}(s_t, a_t) \in \mathbb{R}^{n \times 1}$ with parameters $\omega_{(Ts)}$ is adopted to approximate the action-value $Q^{\pi_\theta}_{Ts}(s_t, a_t)$ of the elements in the transaction scale factor vector $r_t^{Ts}$, which is defined as:

$$Q^{\pi_\theta}_{Ts}(s_t, a_t) \triangleq [q_1^{Ts}, q_2^{Ts}, \ldots, q_n^{Ts}]^T = \mathbb{E}_{s_{i>t}, r_{i \geq t}^{elem}, r_{i \geq t}, a_{i>t} \sim \pi_\theta} \left[ \sum_{i=t}^{\mathcal{T}} \gamma^{i-t} r_t^{Ts} | s_t, a_t \right],$$

where $r_t^{Ts}$ is the transaction scale factor vector defined in Equation (9), and $\gamma$ is as defined in Equation (10). Here, the $i^{th}$ element in the output of the critic network $Q_{\omega_{(Ts)}}(s_t, a_t)$ is defined as $Q^i_{\omega_{(Ts)}}(s_t, a_t)$.

### *4.4 Policy optimization process*

We adopt the DDPG algorithm (Lillicrap et al., 2015) to train the critic networks and the policy network (actor) in our framework. Based on the trajectory in the reply buffer, the critic networks in the multi-critic framework all learn based on the one-step Bellman residual (Sutton, 1988). Here, the critic networks' training objective functions are constructed using the SmoothL1 loss function defined by Girshick (2015). Hence, the training objective functions of critic networks defined in the multi-critic framework (i.e., $Q_{\omega_{(Re)}}(s_t, a_t)$, $Q_{\omega_{(Va)}}(s_t, a_t)$, $Q_{\omega_{(Co)}}(s_t, a_t)$ and $Q_{\omega_{(Ts)}}(s_t, a_t)$) are, respectively:

$$\begin{aligned}
\mathcal{L}_{Q_{(Re)}}(\omega_{(Re)}) &= \mathbb{E}_{(s_i, a_i, r_i^{elem}, r_i, s_{i+1}) \sim B} \left[ SmoothL1loss \left[ Q_{\omega_{(Re)}}(s_i, a_i), (r_i^{Re} + \gamma Q_{\omega'_{(Re)}}(s_{i+1}, \pi_{\theta'}(s_{i+1}))) \right] \right] \\
\mathcal{L}_{Q_{(Va)}}(\omega_{(Va)}) &= \mathbb{E}_{(s_i, a_i, r_i^{elem}, r_i, s_{i+1}) \sim B} \left[ SmoothL1loss \left[ Q_{\omega_{(Va)}}(s_i, a_i), (r_i^{Va} + \gamma Q_{\omega'_{(Va)}}(s_{i+1}, \pi_{\theta'}(s_{i+1}))) \right] \right] \\
\mathcal{L}_{Q_{(Co)}}(\omega_{(Co)}) &= \mathbb{E}_{(s_i, a_i, r_i^{elem}, r_i, s_{i+1}) \sim B} \left[ SmoothL1loss \left[ Q_{\omega_{(Co)}}(s_i, a_i), (r_i^{Co} + \gamma Q_{\omega'_{(Co)}}(s_{i+1}, \pi_{\theta'}(s_{i+1}))) \right] \right] \\
\mathcal{L}_{Q_{(Ts)}}(\omega_{(Ts)}) &= \mathbb{E}_{(s_i, a_i, r_i^{elem}, r_i, s_{i+1}) \sim B} \left[ SmoothL1loss \left[ Q_{\omega_{(Ts)}}(s_i, a_i), (r_i^{Ts} + \gamma Q_{\omega'_{(Ts)}}(s_{i+1}, \pi_{\theta'}(s_{i+1}))) \right] \right]
\end{aligned} \quad (11)$$

where $\theta'$ and $\omega'$ represent the target network of the actor and critic in the DDPG algorithm. In Equation (11), $\mathbb{E}$ represents the expected value function.

Before defining the training objective function of the policy network, we construct a risk constraint term $\Pi(\theta)$ for the policy function based on the SmoothL1 loss function:

$$\Pi(\theta) = \mathbb{E}_{(s_i, a_i, r_i^{elem}, r_i, s_{i+1}) \sim B} \left[ SmoothL1loss \left[ \mathcal{H} \left[ Q_{\omega_{(Re)}}(s_i, \pi_\theta(s_i)) - \xi \otimes \left[ Q_{\omega_{(Va)}}(s_i, \pi_\theta(s_i)) + Q_{\omega_{(Co)}}(s_i, \pi_\theta(s_i)) \right] \right], 0 \right] \right], \quad (12)$$

where $\otimes$ denotes the element-wise multiple, and $\mathcal{H}[\cdot]$ is a piecewise function, which is based on the ReLU activation function:

$$\mathcal{H}(x) = -\text{ReLU}(-x).$$

In Equation (12), $\xi \in \mathbb{R}^{n \times 1}$ contains the risk aversion of each asset in the portfolio, which is defined as

$$\xi = [\xi_1, \xi_2, \ldots, \xi_n]^T,$$

where the $i^{th}$ element $\xi_i$ represents the investor's risk aversion of the $i^{th}$ asset in the portfolio. By adding the risk constraint term $\Pi(\theta)$ in the training objective function of the policy $\pi_\theta$, our training target is to constrain the action-value of the return $Q^i_{\omega_{(Re)}}(s_i, \pi_\theta(s_i))$ modified by the action-value of the variance $Q^i_{\omega_{(Va)}}(s_i, \pi_\theta(s_i))$ and covariance $Q^i_{\omega_{(Co)}}(s_i, \pi_\theta(s_i))$ of each portfolio asset $i$ not smaller than zeros:

$$Q^i_{\omega_{(Re)}}(s_i, \pi_\theta(s_i)) - \xi_i \left[ Q^i_{\omega_{(Va)}}(s_i, \pi_\theta(s_i)) + Q^i_{\omega_{(Co)}}(s_i, \pi_\theta(s_i)) \right] \geq 0, \; for \; i = 1, 2, \ldots, N.$$

In this way, we further reduce the action space that needs to be explored and improve the efficiency of optimal policy exploration.

As described in Figure 3, the training objective function of the policy network is constructed based on the critic networks defined in Section 4.3 and the risk constraint term $\Pi(\theta)$ defined in Equation (12):

$$\mathcal{L}_\pi(\theta) = \mathbb{E}_{(s_i, a_i, r_i^{elem}, r_i, s_{i+1}) \sim B} \left[ \frac{1}{n} \sum_{i=1}^{n} \left[ Q^i_{\omega_{(Re)}}(s_i, \pi_\theta(s_i)) - (\lambda_1/100) \left[ Q^i_{\omega_{(Va)}}(s_i, \pi_\theta(s_i)) + Q^i_{\omega_{(Co)}}(s_i, \pi_\theta(s_i)) \right] - (\lambda_2 * 100) \left[ Q^i_{\omega_{(Ts)}}(s_i, \pi_\theta(s_i)) \right] \right] \right] + \lambda_3 \Pi(\theta), \quad (13)$$

where $n$ is the number of assets in the portfolio. In Equation (13), the parameters $\lambda_1$ and $\lambda_2$ are as defined in Equation (5). Since the elements in the reward factor matrix are calculated based on percentage-based return data, the calibration parameters for the terms $\left[ Q^i_{\omega_{(Va)}}(s_i, \pi_\theta(s_i)) + Q^i_{\omega_{(Co)}}(s_i, \pi_\theta(s_i)) \right]$ and $\left[ Q^i_{\omega_{(Ts)}}(s_i, \pi_\theta(s_i)) \right]$ should be, respectively, reduced to 1% of and 100 times the original values defined in Equation (5).

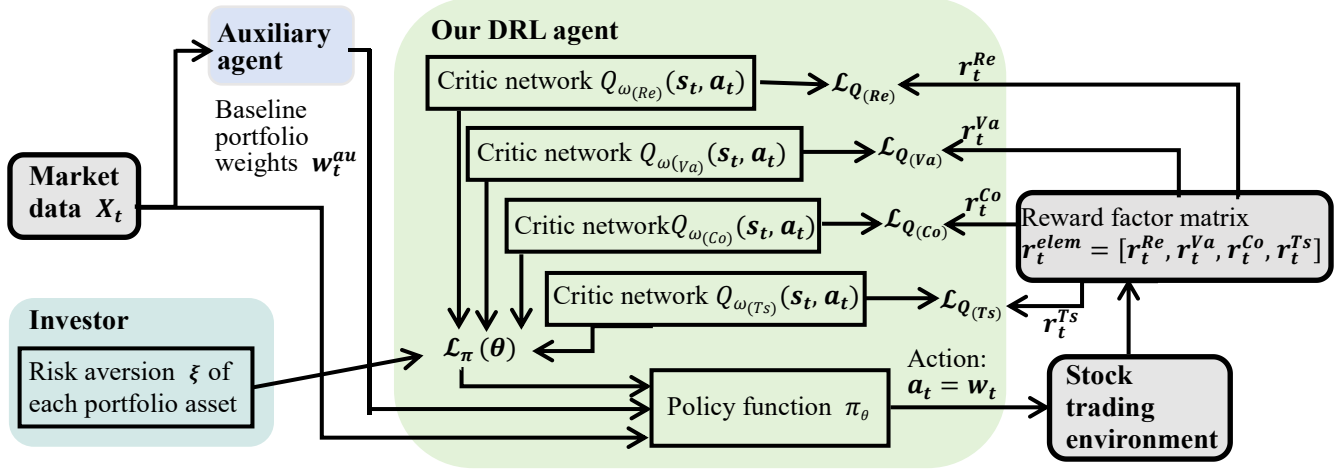

**Fig 3:** The illustration of the multi-critic framework in the training of our DRL agent. In the framework, four critic networks are adopted to learn four corresponding constituted factor vectors in the reward factor matrix $r_t^{elem}$. In the training process of the policy function, the training objective function $\mathcal{L}_\pi(\theta)$ is constructed based on these four critic networks and the investor's risk aversion of each portfolio asset.

## 5. Experimental results
### 5.1 Data Description and Experiment Setting

In our research, we conduct experiments based on historical data to examine the training effectiveness of the current algorithm on our Factor-MCLS and evaluate its performance in out-of-sample profitability and risk control ability. We test our Factor-MCLS in the U.S. stock market to meet the assumption A3 proposed in Section 2.2. Given that the U.S. stock market allows short selling and is the world's most influential stock market, it serves as the ideal experimental market for our Factor-MCLS.

To satisfy the assumptions A1 and A2 specified in Section 2.2, the stocks included in our portfolio must exhibit robust liquidity. Therefore, we include the constituent stocks of the Dow Jones Industrial Average (DJIA) in our portfolio. DJIA consists of thirty of the largest U.S. companies by market capitalization. Hence, the constituent stocks of the DJIA typically demonstrate high levels of liquidity, characterized by substantial market capitalization and significant trading volumes. Following the removal of stocks with missing data, a total of 29 constituent stocks have been identified in our analysis. The stock data used covers the period from January 2019 to December 2022 and has been sourced from Yahoo Finance[1].

To assess the out-of-sample performance of our Factor-MCLS in different market scenarios, we conduct two distinct back-testing experiments under different market scenarios. Specifically, during the back-testing period of Expt 1, the DJIA exhibits oscillating in a downside channel. In contrast, during the back-testing period of Expt 2, the DJIA experiences a fluctuating upward trend. In each experiment, we track multiple key indices which reflect the improvement of our Factor-MCLS's profitability and risk control ability during the training process. Meanwhile, we comprehensively evaluate the profit and risk control abilities of our Factor-MCLS in each back-testing experiment. Table 1 below presents the time ranges of the training and back-testing datasets for each experiment. As shown in Table 1, in each experiment, the training environment is constructed based on three years of historical data. After the DRL agent completes training in the training environment, a back-test is conducted over the subsequent 120 trading days. The indices used in the training and back-testing experiments are detailed in Section 5.2 and Section 5.5.

| Experiment | Training set | Back-testing set |
|---|---|---|
| Expt 1 | 2019.01.01 - 2021.12.31 | 120 trading days starting from 2022.01.01 |
| Expt 2 | 2019.07.01 - 2022.06.30 | 120 trading days starting from 2022.07.01 |

**Table 1:** The time horizon of the training set and the back-testing set across different experiments.

### 5.2 Significant Indices Tracking in the Training Process

In the training process, we track the changes in some significant indices in the training environment. Upon completing each stage of training, we record the value of each term in the objective function of our actor-critic algorithm. Simultaneously, we evaluate our DRL agents in the training environment and record significant indices reflecting the profitability and risk control ability of the DRL agents. During the evaluation process, our agents determine the target portfolio weights $w_t$ and reallocate the assets in each trading period of the training environment. Based on the changes in the total assets, we calculate the logarithm return $r_t^{ta}$ of the investment at the end of each trading period. Furthermore, to measure the risk undertaken in the investment, we calculate the potential variance $\varepsilon_t^{ta}$ of the target portfolio $w_t$ determined by our DRL agent. The calculations of the relative return $r_t^{ta}$ and the potential variance $\varepsilon_t^{ta}$ of the investment in the $t^{th}$ trading period of the training environment are given as follows:

---
[1] http://www. finance.yahoo.com

$$r_t^{ta} = \left(\frac{v_{t_K}^{total}}{v_{(t-1)_K}^{total}}\right) - 1, \quad (14)$$
$$\varepsilon_t^{ta} = \boldsymbol{w}_t^T \boldsymbol{\Sigma}_t \boldsymbol{w}_t$$

where $v_{t_K}^{total}$ is the value of total assets at the end of the last trading day $K$ within the $t^{th}$ trading period, and the term $\boldsymbol{w}_t^T \boldsymbol{\Sigma}_t \boldsymbol{w}_t$ is the variance of the portfolio as defined in Equation (5).

The tracked indices can be divided into two kinds: the indices reflecting the profitability, risk control ability of our DRL agent in the training environment, and the indices reflecting the value of each term in the training objective function. The trajectories of the tracked indices are given in Figure 4 and Figure 5. Here, the tracked indices, which reflect the profitability and risk control ability of our DRL agent, are all calculated based on the logarithmic return $r_t^{ta}$ and the potential variance $\varepsilon_t^{ta}$ defined in Equation (14). The specific definitions of the tracked indices are given as follow:

**The accumulated return $AR^{(tra)}$** is the total daily logarithmic return value obtained by our DRL agent in the training environment, which is a significant index in describing the profitability of our DRL agent:

$$AR^{(tra)} = \left[\prod_{t=1}^{T_f^{tr}} [r_t^{ta} + 1]\right] - 1, \quad (15)$$

where $r_t^{ta}$ is the logarithmic return defined in Equation (14), and $T_f^{tr}$ is the number of trading periods in the training environment. The trajectory of such an index is given on the left-hand side of Figure 4(a).

**The accumulated reward $ARD^{(tra)}$** is the total reward value obtained by our DRL agent in the training environment. Since the calculation of the reward is defined as the return modified by the corresponding variance and transaction scale, the accumulated return could reflect the DRL agent's performance in profitability and risk control ability simultaneously:

$$ARD^{(tra)} = \sum_{t=1}^{T_f^{tr}} r_t, \quad (16)$$

where $r_t$ is the reward obtained by our DRL agent in the $t^{th}$ trading period (Equation (5)). The trajectory of such an index is given on the right-hand side of Figure 4(a).

**The accumulated variance $AV^{(tra)}$** is the total variance value of the portfolio weights determined by our DRL agent in the training environment; the index is adopted to describe the risk taken by our DRL agent:

$$AV^{(tra)} = \sum_{t=1}^{T_f^{tr}} \varepsilon_t^{ta},$$

where $\varepsilon_t^{ta}$ is the variance of the portfolio as defined in Equation (14), and $T_f^{tr}$ is as defined in Equation (15). The trajectory of such an index is given in Figure 4(c).

**Number of periods with positive return ($NPR^{(tra)}$) and number of periods with positive reward ($NPRW^{(tra)}$).** To describe the returns and rewards our DRL agent obtained in the training environment from a different perspective, we adopt two indices: the number of days with positive return (NDP) and the number of days with positive reward (NDPR) in the indices tracking in the training process. The number of days with positive return (NDP) is defined based on the indicator function $I(\cdot)$:

$$NPR^{(tra)} = \sum_{t=1}^{T_f^{tr}} I(r_t^{ta} > 0),$$

where $T_f^{tr}$ and $r_t^{ta}$ are as defined in Equation (15). The number of periods with positive reward (NDPR) is defined as:

$$NPRW^{(tra)} = \sum_{t=1}^{T_f^{tr}} I(r_t > 0),$$

where $r_t$ is as defined in Equation (5). The trajectories of NPR and NPRW are both given in Figure 4(d).

To demonstrate that the current training algorithm is resistant to gradient vanishing and gradient exploding, we track the values of the training objective functions and their components throughout the training process. The specific definitions of the tracked indices, which are related to the training objective functions in our actor-critic algorithm, are given as follows:

**The value of the training objective functions.** At first, we track the training objective function of the policy function without the risk constraint term $\mathcal{L}_\pi^{wr}(\theta)$:

$$\mathcal{L}_\pi^{wr}(\theta) = \mathbb{E}_{(s_i, a_i, r_i^{elem}, r_i, s_{i+1}) \sim B} \left[\frac{1}{N}\sum_{i=1}^N \left[Q_{\omega_{(Re)}}^i(s_i, \pi_\theta(s_i)) - (\lambda_1/100)\left[Q_{\omega_{(Va)}}^i(s_i, \pi_\theta(s_i)) + Q_{\omega_{(Co)}}^i(s_i, \pi_\theta(s_i))\right]\right.\right. \\ \left.\left. - (\lambda_2 * 100)\, Q_{\omega_{(Ts)}}^i(s_i, \pi_\theta(s_i))\right]\right], \quad (17)$$

and the sum of the training objective functions of the critic networks in our multi-critic framework:

$$\mathcal{L}_Q^{total}(\omega) = \mathcal{L}_{Q_{(Re)}}(\omega_{(Re)}) + \mathcal{L}_{Q_{(Va)}}(\omega_{(Va)}) + \mathcal{L}_{Q_{(Co)}}(\omega_{(Co)}) + \mathcal{L}_{Q_{(Ts)}}(\omega_{(Ts)}). \quad (18)$$

The trajectories of these two indices are, respectively, given on the left-hand side and right-hand side of Figure 4(b). Furthermore, we track the specific value of each critic network's training objective function. The trajectories of these four indices are given on the right-hand side of Figure 5(a)-5(d).

**The value of each term in the training objective functions.** Except the indices defined in Equation (17) and (18), we respectively track the specific value of each term in the training objective function defined in Equation (13):

$$\mathcal{L}_\pi^{Re}(\theta) = \mathbb{E}_{(s_i, a_i, r_i^{elem}, r_i, s_{i+1}) \sim B} \left[\frac{1}{N}\sum_{i=1}^N Q_{\omega_{(Re)}}^i(s_i, \pi_\theta(s_i))\right]$$
$$\mathcal{L}_\pi^{Va}(\theta) = \mathbb{E}_{(s_i, a_i, r_i^{elem}, r_i, s_{i+1}) \sim B} \left[\frac{1}{N}\sum_{i=1}^N Q_{\omega_{(Va)}}^i(s_i, \pi_\theta(s_i))\right]$$
$$\mathcal{L}_\pi^{Co}(\theta) = \mathbb{E}_{(s_i, a_i, r_i^{elem}, r_i, s_{i+1}) \sim B} \left[\frac{1}{N}\sum_{i=1}^N Q_{\omega_{(Co)}}^i(s_i, \pi_\theta(s_i))\right]$$
$$\mathcal{L}_\pi^{Ts}(\theta) = \mathbb{E}_{(s_i, a_i, r_i^{elem}, r_i, s_{i+1}) \sim B} \left[\frac{1}{N}\sum_{i=1}^N Q_{\omega_{(Ts)}}^i(s_i, \pi_\theta(s_i))\right]$$

$$\Pi(\theta) = \mathbb{E}_{(s_i, a_i, r_i^{elem}, r_i, s_{i+1}) \sim B} \left[SmoothL1loss\left[\mathcal{H}\left[Q_{\omega_{(Re)}}(s_i, \pi_\theta(s_i))\right] - \xi \otimes \left[Q_{\omega_{(Va)}}(s_i, \pi_\theta(s_i)) + Q_{\omega_{(Va)}}(s_i, \pi_\theta(s_i))\right]\right], 0\right]\right]$$

The trajectory of the terms $\mathcal{L}_\pi^{Re}(\theta), \mathcal{L}_\pi^{Va}(\theta), \mathcal{L}_\pi^{Ts}(\theta)$, and $\mathcal{L}_\pi^{Co}(\theta)$ are, respectively, given in the left-hand side of Figure 5(b)-5(e). The trajectory of the risk constraint term $\Pi(\theta)$ is given on the right-hand side of Figure 5(f). Meanwhile, we track the value of the training objective function of each critic network in the multi-critic framework defined in Equation (11), and the trajectory of these four indices are, respectively, given on the right-hand side of Figure 5(b)-5(e).

**The value of our policy function.** Furthermore, to evaluate the policy of our DRL agent based on the action value of the reward function defined in Equation (5), we train a new critic network $Q_\varphi(s_t, a_t)$ to learn the value function:

$$\mathbb{E}_{s_{i>t}, r_{i\geq t}, a_{i>t} \sim \pi_\theta} \left[ \sum_{i=t}^{T} \gamma^{i-t} r_t | s_t, a_t \right],$$

where the reward $r_t$ is calculated based on the reward function defined in Equation (5). Hence, the training objective function of the critic network $Q_\varphi(s_t, a_t)$ is defined as:

$$\mathcal{L}(\varphi) = \mathbb{E}_{(s_i, a_i, r_i^{elem}, r_i, s_{i+1}) \sim B} \left[ SmoothL1loss[Q_\varphi(s_i, a_i), r_i + \gamma Q_{\varphi'}(s_{i+1}, \pi_{\theta'}(s_{i+1}))] \right].$$

Based on the critic network $Q_\varphi(s_t, a_t)$, we track the critic value of our policy function in the training process:

$$\mathbb{E}_{(s_i, a_i, r_i^{elem}, r_i, s_{i+1}) \sim B} [Q_\varphi(s_i, \pi_\theta(s_i))]. \tag{19}$$

The trajectory of the index defined in Equation (19) is given on the right-hand side of Figure 5(f).

## 5.3 Performance in the training environment

In this section, we conduct a trajectory analysis of the indices proposed in section 5.2. The indices included in the analysis are categorized into two groups: those from the training environment and those related to the training objective functions. The trajectories of the indices from the training environment are presented in Figure 4(a), 4(c), and 4(d). The trajectories of the indices pertaining to the training objective functions are shown in Figure 4(b) and Figure 5(a) to Figure 5(f).

From the trajectories of the indices given in Figure 4(a), we find that the accumulated return and the accumulated reward obtained by our Factor-MCLS are significantly improved in the training process. Since the reward is a risk-adjusted return index, we may conclude that our Factor-MCLS's profitability and risk-adjusted profitability are significantly improved in the training. In Figure 4(c), we observe that the accumulated variance taken by our Factor-MCLS is kept at a low level. This indicates that it can learn to effectively control the portfolio's risk in the training environment. From the trajectories of the number of periods with positive returns and the number of periods with positive rewards given in Figure 4(d), we report that our Factor-MCLS can consistently achieve positive returns in each trading period within the training environment during the training process. This suggests that by utilizing an auxiliary agent, it can concentrate on exploring a policy that yields higher rewards within the action space that generates positive returns. Furthermore, the trajectory of the number of periods with positive rewards indicates that our Factor-MCLS can attain positive rewards in at least 95% of the trading periods within the training environment throughout the training process. This indicates that the currently adopted training objective function can effectively ensure that it can focus on exploring policy within the action space that receives positive rewards for most trading periods.

In addition to conducting trajectory analysis on profitability and risk control ability indices within the training environment, we also perform a convergence analysis of the objective functions employed for training the DRL agent and its various components. The Figure 4(b), as well as Figure 5(b)-5(e), illustrate the value trajectories of the actor's training objective function and the objective function's four components $\mathcal{L}_\pi^{Re}(\theta)$, $\mathcal{L}_\pi^{Va}(\theta)$, $\mathcal{L}_\pi^{Ts}(\theta)$, and $\mathcal{L}_\pi^{Co}(\theta)$ throughout the training process. The trajectory plots indicate that the actor's training objective function and its components $\mathcal{L}_\pi^{Re}(\theta)$, $\mathcal{L}_\pi^{Va}(\theta)$, $\mathcal{L}_\pi^{Ts}(\theta)$, and $\mathcal{L}_\pi^{Co}(\theta)$ achieve stable convergence during the training process, without any issues of gradient explosion. Figure 4(b), and the right-hand sides of Figure 5(b)-5(e) illustrate the value trajectories of the training objective functions for each critic network and their sum throughout the training process. The trajectory plots indicate that the loss function for training the critic network demonstrates a stable tendency to converge to zero throughout the training process without gradient explosion. Based on the trajectories of accumulated return and accumulated reward depicted in Figure 4(a), we conclude that the critic network is able to effectively learn from the information contained within the reward factor matrix $r_{elem}$. Additionally, through gradient propagation, the critic network facilitates the effective training of the actor network, thereby enhancing both the profitability and the risk-adjusted profitability of the learning system. Additionally, Figure 5(f) shows that the value of the risk constraint term demonstrates a stable tendency to converge to zero. This observation and the effective control of accumulated variance during training, as demonstrated in Figure 4(c), indicate successful risk management. This suggests that the risk constraint term in the actor's training objective function effectively enhances the DRL agent's risk control ability. To evaluate the actions of the learning system from the perspective of long-term rewards, we depict a new critic network that learns rewards directly from the environment and provides Q-value estimates for the actions output by our Factor-MCLS. The Figure 5(f) illustrates the trajectory of Q-value estimations for the actions generated by our Factor-MCLS in the training process. The trajectory of Q-value changes reveals that the Q-values for the actions produced by the actor of our DRL agent converge smoothly after a phase of steady growth during training.

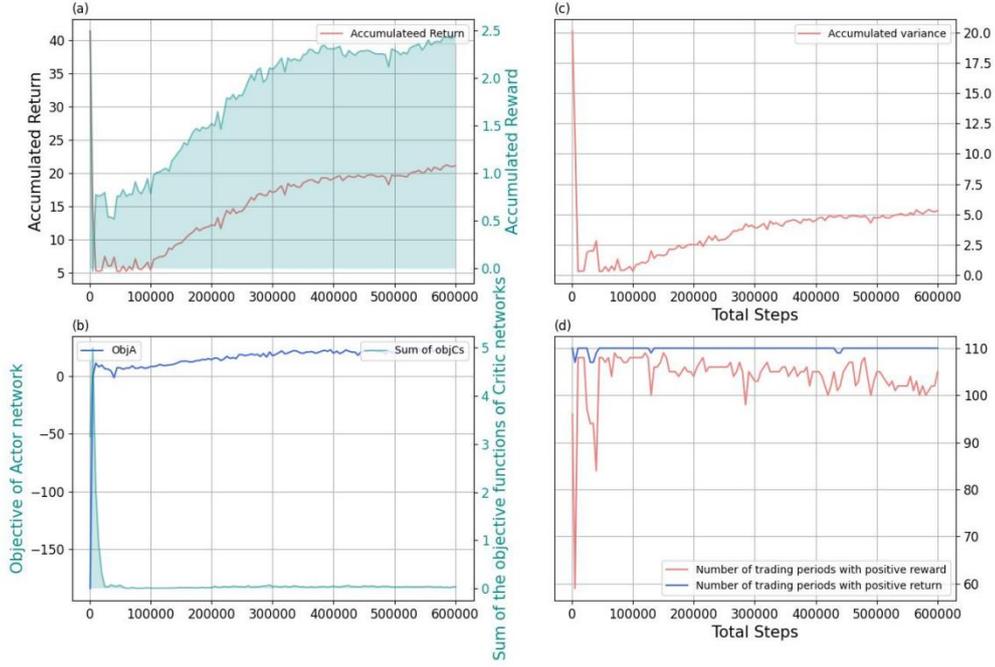

**Fig 4:** Significant indices are tracked in the training process. During the training process, we monitor several indices associated with both the training environment and the training objective functions, with their specific trajectories illustrated in subplot (a)-(d). Subplot (a) presents the trajectories of the accumulated return and accumulated reward obtained by our Factor-MCLS throughout the training process. Subplot (b) displays the trajectories of the training objective function for the actor network, along with the total value of the training objective functions for all critic networks. Subplot (c) illustrates the trajectories of the accumulated variance. Finally, subplot (d) depicts the trajectories of the number of periods with positive rewards and the number of periods with positive returns.

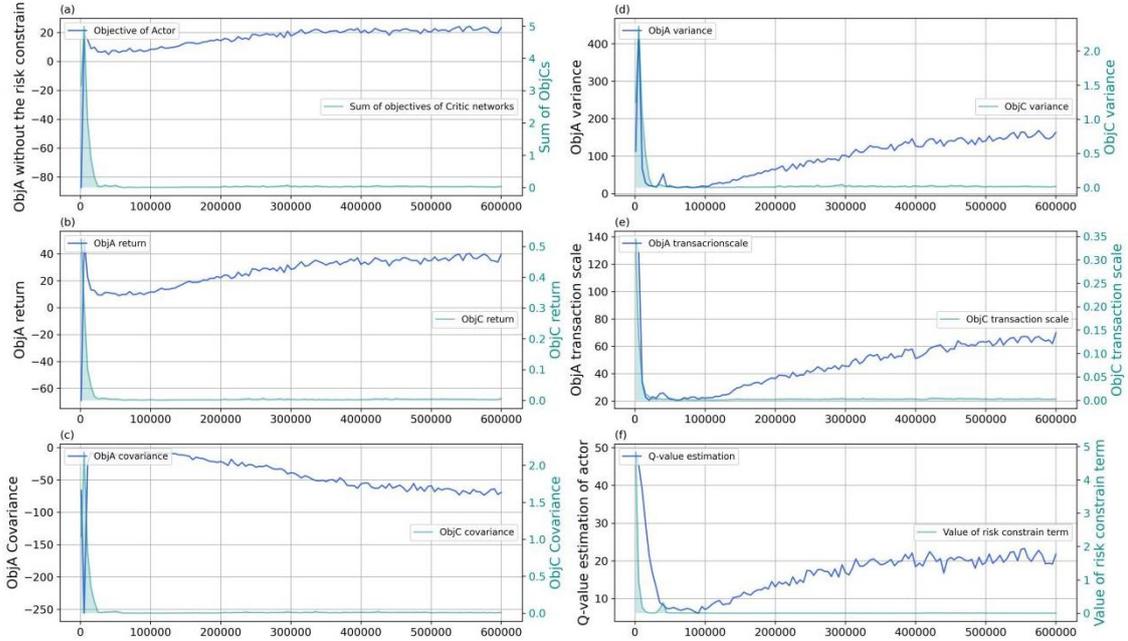

**Fig 5:** The specific values of the terms in the training objective functions. The training objective functions, whose trajectories during the training process are illustrated in subplot (b) of Figure 4, include several terms related to return, variance, covariance, and transaction scale. We present the trajectories of each term within these training objective functions. In our multi-critic framework, we utilize four critic networks, each with its own training objective function, whose trajectories in the training process are displayed on the right-hand side of subplots (b) to (e). The trajectories of the terms $\mathcal{L}_\pi^{Re}(\theta)$, $\mathcal{L}_\pi^{Va}(\theta)$, $\mathcal{L}_\pi^{Tr}(\theta)$, and $\mathcal{L}_\pi^{Co}(\theta)$ in the training objective function of the actor network are presented on the left-hand side of subplots (b) to (e). Furthermore, the trajectory of the risk constraint term is shown on the right-hand side of the subplot (f).

Additionally, the trajectory of the training objective function of the actor, excluding the risk constraint term defined in Equation (17), is depicted on the left-hand side of subplot (a). The trajectory of the index defined in Equation (19) is represented on the left-hand side of subplot (f).

### 5.4 Comparative Portfolio Rules in the Back-tests

To demonstrate the outstanding out-of-sample performance of our Factor-MCLS in profitability and risk management, it is compared with the following dynamic portfolio optimization strategies proposed in recent years in back-test experiments:

**Capital growth theory (Kelly, 1956).** The comparative strategies based on the capital growth theory can be classified into five major classifications, namely baseline strategies, follow-the-winner, follow-the-loser, pattern-matching approaches, and meta-learning algorithms. Here, we employ 14 dynamic portfolio strategies belonging to these five classifications as comparative benchmarks.

**Machine learning algorithms.** The comparative strategies leveraging machine learning algorithms are categorized into two distinct groups: those utilizing DL-based algorithms and those employing DRL-based algorithms. Within the DL-based strategies, dynamic portfolio optimization strategies are formulated following the approach delineated by Duan et al. (2022) in their research. Specifically, TopK-Drop strategies[2] are implemented to determine portfolio weights based on DL model predictions. In the DRL-based strategies, we adopt five DRL comparative strategies sourced from FinRL (Liu et al., 2021), an open-source library dedicated to financial reinforcement learning. Within the FinRL framework, Liu et al. (2021) have constructed a comprehensive pipeline tailored for deploying DRL algorithms in financial portfolio management. This pipeline incorporates cutting-edge DRL methodologies fine-tuned for financial contexts. Furthermore, we adopt the strategies named EIIE of Jiang et al. (2017) as the comparative strategy. As a deep policy gradient algorithm, EIIE (Jiang et al., 2017) is supplementary to the strategies in the FinRL library. A comprehensive classification of these comparative strategies is provided in Table 1.

| Categories | Classifications | Algorithm |
|---|---|---|
| Strategies based on Capital Growth Theory | Baseline strategies | Constant Rebalanced Portfolios (CRP) (Cover, 1991) |
| | | M0 (M0) (Borodin et al., 2000) |
| | | Uniform Buy And Hold (UBAH) (Li & Hoi, 2014) |
| | Follow-the-Winner | Universal Portfolio (UP) (Cover, 1991; Cover, 1996) |
| | | Exponentiated Gradient (EG) (Helmbold et al., 1998) |
| | Follow-the-Loser | Anti-Correlation (ANTICOR) (Borodin et al., 2003) |
| | | Passive Aggressive Mean Reversion (PAMR) (Li et al., 2012) |
| | | Confidence Weights Mean Reversion (CWMR) (Li et al., 2011) |
| | | On-Line Portfolio Selection with Moving Average Reversion (OLMAR) (Li & Hoi, 2012) |
| | | Robust Median Reversion (RMR) (Huang et al., 2012) |
| | | Weighted Moving Average Mean Reversion (WMAMR) (Gao & Zhang, 2013) |
| | Pattern-Matching Approaches | Nonparametric Kernel-Based Log Optimal Strategy (BK) (Györfi, 2006) |
| | | Correlation-driven Nonparametric learning (CORN) (Li et al., 2011) |
| | Meta-Learning Algorithm | Online Newton Step (ONS) (Agarwal et al., 2006) |
| Strategies based on machine learning algorithms | DRL-based strategies | EIIE (Jiang et al., 2017) |
| | | DDPG-FinRL (Liu et al., 2021) |
| | | SAC-FinRL (Liu et al., 2021) |
| | | PPO-FinRL (Liu et al., 2021) |
| | | TD3-FinRL (Liu et al., 2021) |
| | | A2C-FinRL (Liu et al., 2021) |
| | | BDA (Sun et al., 2024) |
| | DL-based strategies | Dlinear (Zeng et al., 2023) |
| | | Autoformer (Wu et al., 2021) |
| | | Informer (Zhou et al., 2021) |
| | | PatchTST (Nie et al., 2022) |

**Table 1:** Comparative strategies. The table presents a comprehensive overview of the comparative strategies implemented in each back-testing experiment.

### 5.5 Back-test Performance Measure

In the back-test experiments, we calculate the logarithm return $r_{t_k}^{bt}$ of the investment at each trading day. According to the trajectories of the daily logarithm returns in the back-test, we adopt several indices to measure the performance of our DRL agent. In these indices, accumulated return (AR) and daily return (DR) are adopted to measure the profitability of our DRL agent and the comparative strategies. At the same time, standard deviation (Std) and low partial standard deviation (LStd) are adopted to measure the risk control ability. Furthermore, to comprehensively evaluate the ability of the DRL agent to attain profit and control risk, we adopt the Sharpe ratio (SR) and Sortino ratio (STR) to assess the return per unit of risk. The detailed definitions of these indices are given as follows.

---

[2] https://qlib.readthedocs.io/en/latest/component/strategy.html

**Daily logarithmic return in the back-test** ($r_{t_k}^{bt}$). In the back-test, the logarithmic return of the total assets at the $k^{\text{th}}$ trading day in the $t^{\text{th}}$ trading period is defined as:

$$r_{t_k}^{bt} = \begin{cases} \log_2\left(\frac{v_{t_k}}{v_{t_{k-1}}}\right) & k = 2, 3, \ldots, K \\ \log_2\left(\frac{v_{t_k}}{v_{t-1_K}}\right) & k = 1 \end{cases}, \quad (20)$$

where $v_{t_k}$ is the value of total assets at the end of the $k^{\text{th}}$ trading day in the $t^{\text{th}}$ trading period.

**Accumulated return** (AR). In the same way as the calculation of the accumulated return in the training environment, we also calculate the accumulated return in the back-test based on the daily return $r_{t_k}^{bt}$ defined in Equation (20):

$$\text{AR} = \sum_{t=1}^{T_f^{bt}} \sum_{k=1}^{K} r_{t_k}^{bt}, \quad (21)$$

where $T_f^{bt}$ is the number of trading periods in the back-test environment, and $K$ is the number of trading days in a trading period.

**Daily return (DR).** Daily return is the mean of the logarithmic daily return in the back-test, which is defined as:

$$\text{DR} = \frac{1}{T_f^{bt} K} \sum_{t=1}^{T_f^{bt}} \sum_{k=1}^{K} r_{t_k}^{bt}, \quad (22)$$

where $T_f^{bt}$ and $K$ are as defined in Equation (21).

**Variance** (Var) **and standard deviation** (Std). In our experiments in the back-tests, we adopt the variance and standard deviation to measure the risk control ability of our DRL agent. The indices are respectively defined as:

$$\text{Var} = \frac{1}{T_f^{bt} K} \sum_{t=1}^{T_f^{bt}} \sum_{k=1}^{K} \left[r_{t_k}^{bt} - \frac{1}{T_f^{bt} K} \sum_{t=1}^{T_f^{bt}} \sum_{k=1}^{K} r_{t_k}^{bt}\right]^2$$

$$\text{Std} = \sqrt{\frac{1}{T_f^{bt} K} \sum_{t=1}^{T_f^{bt}} \sum_{k=1}^{K} \left[r_{t_k}^{bt} - DR\right]^2}, \quad (23)$$

where $DR$ is the daily return defined in Equation (22), and the indices $T_f^{bt}, K$, and $r_{t_k}^{bt}$ are as defined in Equation (21).

**Low partial standard deviation** (LStd). Drawing upon Rollinger and Hoffman's foundational perspective (2013), the volatility caused by the increase in the asset price should not be considered risk. Hence, we adopt the low partial standard deviation to measure downside risk:

$$\text{LStd} = \sqrt{\frac{1}{T_f^{bt} K} \sum_{t=1}^{T_f^{bt}} \sum_{k=1}^{K} \left[\min(r_{t_k}^{bt} - M_{ac}, 0)\right]^2}, \quad (24)$$

where $M_{ac}$ is the minimum acceptable return. In Equation (24), $T_f^{bt}, K$, and $r_{t_k}^{bt}$ are as defined in Equation (21).

**Sharpe ratio** (SR). To comprehensively assess the performance of our DRL agents in profitability and risk control, we adopt the Sharpe ratio (Sharpe, 1966) to describe the DRL agent's profit obtained in assuming per unit of risk undertaken. Here, the Sharpe ratio is calculated as:

$$\text{SR} = \frac{\text{DR} - r_f}{\text{Std}}, \quad (25)$$

where $r_f$ is the daily risk-free return. In Equation (25), DR and Std are as defined in Equation (22) and (23).

**Sortino ratio** (STR). Similar to the Sharpe ratio, the Sortino ratio (Rollinger & Hoffman, 2013) is also adopted to measure our DRL agent's profit per unit of risk undertaken. Unlike the the Sharpe ratio, risk is measured by the low partial standard deviation in the calculation of the Sortino ratio. Here, Sortino ratio is calculated as:

$$\text{STR} = \frac{\text{DR} - M}{\text{LStD}},$$

where the indices DR, M, and LStD are as defined in Equation (22) and (24).

## 5.6 *Performance in the back-test experiments*

The performance of Factor-MCLS and comparative strategies in the back-testing experiments is presented in Tables 1 and 2. To offer a clearer representation of the advantages of Factor-MCLS over the comparison strategies in terms of profitability and profitability per unit of risk, we employed the analysis method of Sun et al. (2025) and visualized the data from Tables 1 and 2 in Figures 6-9. Among these four figures, Figures 6 and 8 illustrate the performance of Factor-MCLS and the comparative strategies based on traditional capital growth theory in the back-testing experiments. Figures 7 and 9 illustrate the performance comparison between Factor-MCLS and the machine learning algorithm-based comparative strategy in the back-testing experiments.

Specifically, each figure in Figures 6 through 9 includes four subplots that present the performance of Factor-MCLS and the comparative strategies from various perspectives. The subplot (a) in Figures 6 through 9 illustrates the cumulative return trajectories of Factor-MCLS and several comparative strategies over time. The subplot (b) in Figures 6 through 9 uses a bar chart to visualize the accumulative returns of Factor-MCLS and several comparative strategies in the back-testing experiment. Additionally, we also annotate the cumulative returns for the first 40 and 80 trading days in the bar chart to reflect the return distribution of our Factor-MCLS and the comparative strategies in the back-testing experiments. In subplots (c) and (d) of Figures 6 through 9, the y-axis displays the daily average returns of Factor-MCLS and various comparative strategies in the back-testing experiments. The horizontal axis in subplots (c) and (d) illustrate the standard deviation and the lower partial standard deviation of returns for each trading day within the back-testing experiment. Given that we assume a risk-free return of zero and consider it the minimum acceptable return, the slope of the line connecting each strategy's point to the origin in subplots (c) and (d) represents the Sharpe ratio and Sortino ratio, respectively. Hence, a steeper slope of this line corresponds to a higher Sharpe ratio (or Sortino ratio) for the strategy.

The visualized experimental results presented in subplots (a) and (b) of Figures 6 through 9 indicate that Factor-MCLS

exhibits an advantage in accumulative returns compared to the comparative strategies across various experiments. As shown by the specific data on accumulative return given in Table 2 and Table 3, our Factor-MCLS outperforms the comparative strategies by at least 35.2%. Additionally, the results displayed in subplots (c) and (d) of Figures 6 through 9 demonstrate that Factor-MCLS also maintains an advantage over the benchmark strategies concerning both the Sharpe ratio and Sortino ratio during different phases of the back-testing experiments. As shown by the specific data in Table 2 and Table 3, our Factor-MCLS outperforms the comparison strategy in terms of SR by at least 63.9%. In terms of STR, our Factor-MCLS outperforms the comparative strategies by at least 66.5%. These empirical results indicate that the dynamic portfolio policies learned by Factor-MCLS in the training environment exhibit advantages in both overall profitability and profitability per unit of risk compared to the comparative strategies in the back-testing experiments. Furthermore, this demonstrates that the dynamic portfolio policies learned by Factor-MCLS in the training environment possess outstanding generalization abilities. Particularly during the back-testing phase of the first experiment, when the market is in a volatile downward trend, comparative strategies except BDA proposed by Sun et al. (2024) all incurred losses. In contrast, Factor-MCLS can achieve stable returns. This outcome further demonstrates that the dynamic portfolio policies learned by Factor-MCLS in the training environment exhibit strong generalization capabilities.

|  | AR | DR | Std | SR | LStd | STR |
| --- | --- | --- | --- | --- | --- | --- |
| **Factor-MCLS** | **0.46628** | **0.00389** | 0.02856 | **0.13604** | 0.02601 | **0.22273** |
| BDA | 0.34470 | 0.00287 | 0.03469 | 0.08280 | 0.03012 | 0.13376 |
| BK | -0.12819 | -0.00107 | 0.02255 | -0.04738 | 0.02634 | -0.05991 |
| CRP | -0.18470 | -0.00154 | 0.01804 | -0.08531 | 0.01911 | -0.11030 |
| ONS | -0.19284 | -0.00161 | 0.01917 | -0.08384 | 0.01972 | -0.10904 |
| OLMAR | -0.50575 | -0.00421 | 0.03439 | -0.12255 | 0.03608 | -0.15633 |
| UP | -0.18388 | -0.00153 | 0.01800 | -0.08511 | 0.01907 | -0.11002 |
| Anticor | -0.16548 | -0.00138 | 0.02541 | -0.05426 | 0.02572 | -0.07284 |
| PAMR | -0.61544 | -0.00513 | 0.03478 | -0.14746 | 0.03822 | -0.18373 |
| CORN | -0.15111 | -0.00126 | 0.02383 | -0.05284 | 0.02692 | -0.06787 |
| M0 | -0.23653 | -0.00197 | 0.01966 | -0.10027 | 0.02119 | -0.12636 |
| RMR | -0.67890 | -0.00566 | 0.03470 | -0.16304 | 0.03772 | -0.20074 |
| CWMR | -0.62653 | -0.00522 | 0.03575 | -0.14605 | 0.03879 | -0.18286 |
| EG | -0.17363 | -0.00145 | 0.01669 | -0.08668 | 0.01772 | -0.11098 |
| UBAH | -0.17931 | -0.00149 | 0.01610 | -0.09281 | 0.01728 | -0.11838 |
| WMAMR | -0.19237 | -0.00160 | 0.03294 | -0.04866 | 0.03479 | -0.06516 |
| FinRL-DDPG | -0.27826 | -0.00232 | 0.02104 | -0.11022 | 0.02179 | -0.14348 |
| FinRL-A2C | -0.00682 | -0.00006 | 0.01021 | -0.00557 | 0.01031 | -0.00755 |
| FinRL-PPO | -0.04619 | -0.00038 | 0.01324 | -0.02908 | 0.01396 | -0.03901 |
| FinRL-SAC | -0.21447 | -0.00179 | 0.02162 | -0.08268 | 0.02254 | -0.10612 |
| FinRL-TD3 | -0.17224 | -0.00144 | 0.01890 | -0.07594 | 0.01944 | -0.10112 |
| EIIE | -0.37617 | -0.00313 | 0.03459 | -0.09062 | 0.03539 | -0.11944 |
| Dlinear | -0.25690 | -0.00214 | 0.02268 | -0.09439 | 0.02354 | -0.12453 |
| Autoformer | -0.32720 | -0.00273 | 0.01993 | -0.13684 | 0.02062 | -0.17566 |
| PatchTST | -0.18049 | -0.00150 | 0.01986 | -0.07572 | 0.02125 | -0.10011 |
| Informer | -0.32720 | -0.00273 | 0.01993 | -0.13684 | 0.02062 | -0.17566 |

**Table 2:** Empirical results in experiment 1. The table displays the quantitative outcomes for the indices specified in Section 5.5, comparing Factor-MCLS with various benchmark strategies from back-test experiment 1. Results highlighting superior performance in terms of return and return per unit risk are indicated in bold.

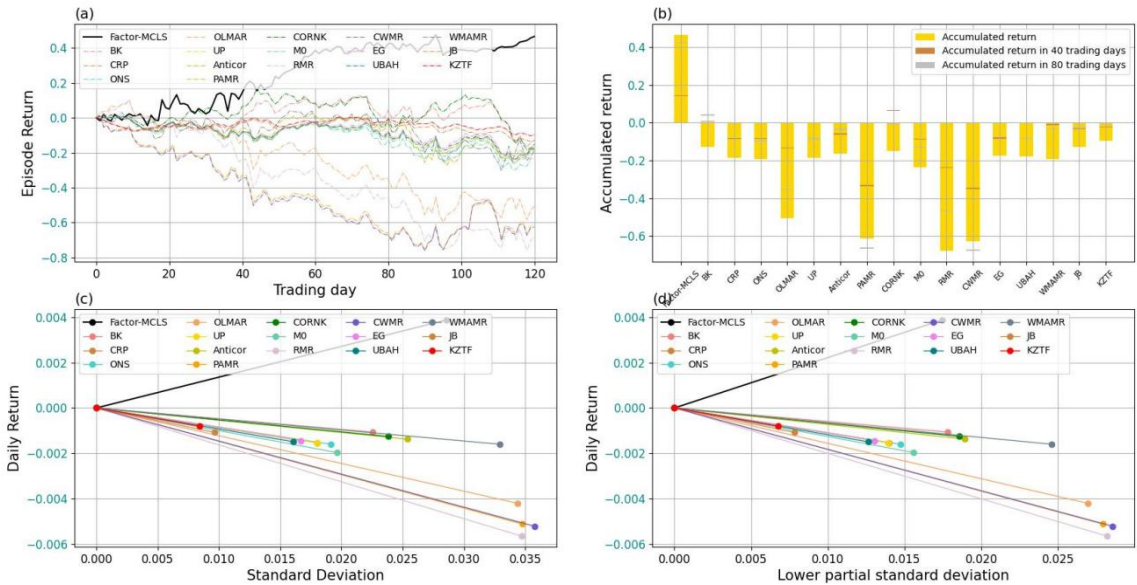

**Fig 6:** Performance Assessment of Portfolio Selection Algorithms in Experiment 1. This figure illustrates the comparative performance of Factor-MCLS against various comparative strategies grounded in Capital Growth Theory and Markowitz's Mean-Variance Theory. The figure visually represents the empirical findings presented in Table 2.

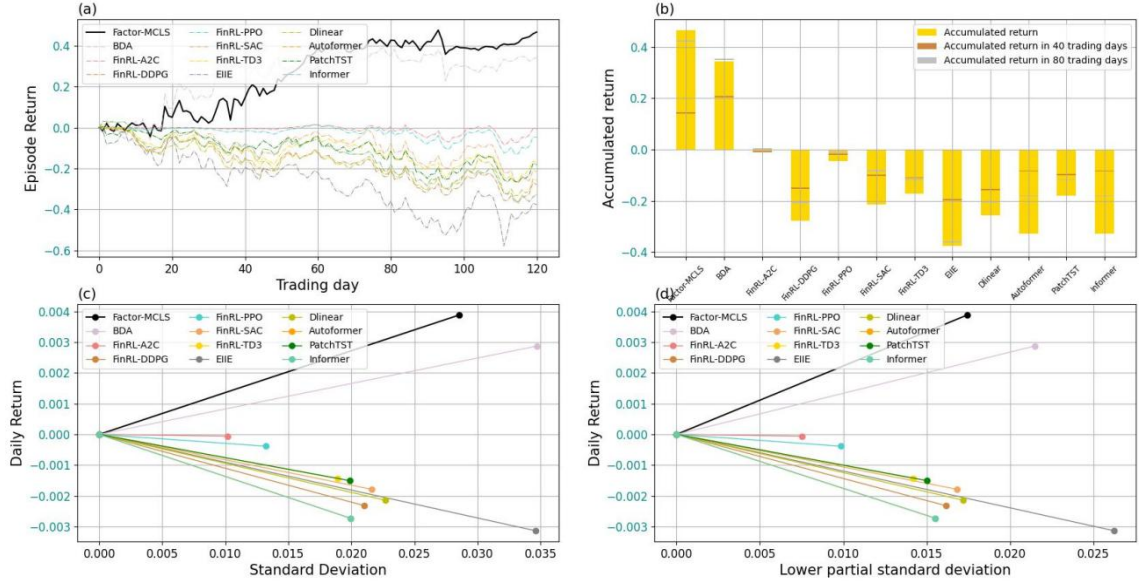

**Fig 7:** Performance Assessment of Portfolio Selection Strategies in Experiment 1. This figure illustrates the comparative performance of Factor-MCLS alongside various comparative strategies based on different machine learning algorithms. It visually represents the empirical results detailed in Table 2.

|  | AR | DR | Std | SR | LStd | STR |
|---|---|---|---|---|---|---|
| **Factor-MCLS** | **0.83419** | **0.00695** | 0.05209 | **0.13347** | 0.04739 | **0.21282** |
| **BDA** | 0.34322 | 0.00286 | 0.03513 | 0.08142 | 0.03518 | 0.11795 |
| **BK** | 0.07825 | 0.00065 | 0.02088 | 0.03123 | 0.02131 | 0.04441 |
| **CRP** | 0.09933 | 0.00083 | 0.01846 | 0.04484 | 0.01782 | 0.06741 |
| **ONS** | 0.09664 | 0.00081 | 0.01861 | 0.04328 | 0.01776 | 0.06580 |
| **OLMAR** | 0.16524 | 0.00138 | 0.03532 | 0.03899 | 0.03175 | 0.06133 |
| **UP** | 0.09947 | 0.00083 | 0.01847 | 0.04488 | 0.01783 | 0.06746 |
| **Anticor** | 0.11395 | 0.00095 | 0.02401 | 0.03954 | 0.02173 | 0.06129 |
| **PAMR** | -0.59267 | -0.00494 | 0.03476 | -0.14208 | 0.03724 | -0.18162 |
| **CORN** | 0.16656 | 0.00139 | 0.02262 | 0.06136 | 0.02250 | 0.09032 |
| **M0** | 0.10031 | 0.00084 | 0.01979 | 0.04223 | 0.01848 | 0.06293 |
| **RMR** | 0.01351 | 0.00011 | 0.03791 | 0.00297 | 0.03729 | 0.00427 |
| **CWMR** | -0.61229 | -0.00510 | 0.03529 | -0.14458 | 0.03760 | -0.18441 |
| **EG** | 0.09751 | 0.00081 | 0.01841 | 0.04414 | 0.01797 | 0.06561 |
| **UBAH** | 0.08677 | 0.00072 | 0.01864 | 0.03879 | 0.01840 | 0.05703 |
| **WMAMR** | 0.09180 | 0.00077 | 0.03268 | 0.02341 | 0.03415 | 0.03251 |
| **FinRL-DDPG** | 0.08345 | 0.00070 | 0.02083 | 0.03338 | 0.01866 | 0.05064 |
| **FinRL-A2C** | 0.03167 | 0.00026 | 0.00952 | 0.02773 | 0.00892 | 0.04219 |
| **FinRL-PPO** | 0.03868 | 0.00032 | 0.01247 | 0.02585 | 0.01153 | 0.03920 |
| **FinRL-SAC** | 0.09615 | 0.00080 | 0.01919 | 0.04175 | 0.01773 | 0.06289 |
| **FinRL-TD3** | 0.06681 | 0.00056 | 0.01443 | 0.03858 | 0.01350 | 0.05693 |
| **EIIE** | 0.24018 | 0.00200 | 0.02551 | 0.07845 | 0.02385 | 0.12511 |
| **Dlinear** | 0.11933 | 0.00099 | 0.02244 | 0.04432 | 0.02217 | 0.06625 |
| **Autoformer** | 0.12359 | 0.00103 | 0.01686 | 0.06109 | 0.01743 | 0.08890 |
| **PatchTST** | 0.17349 | 0.00145 | 0.02027 | 0.07133 | 0.01895 | 0.11069 |
| **Informer** | 0.12359 | 0.00103 | 0.01686 | 0.06109 | 0.01743 | 0.08890 |

**Table 3:** Empirical results in experiment 2. The table displays the quantitative outcomes for the indices specified in Section 5.5, comparing Factor-MCLS with various benchmark strategies from back-test experiment 2. Results highlighting superior performance in terms of return and return per unit risk are indicated in bold.

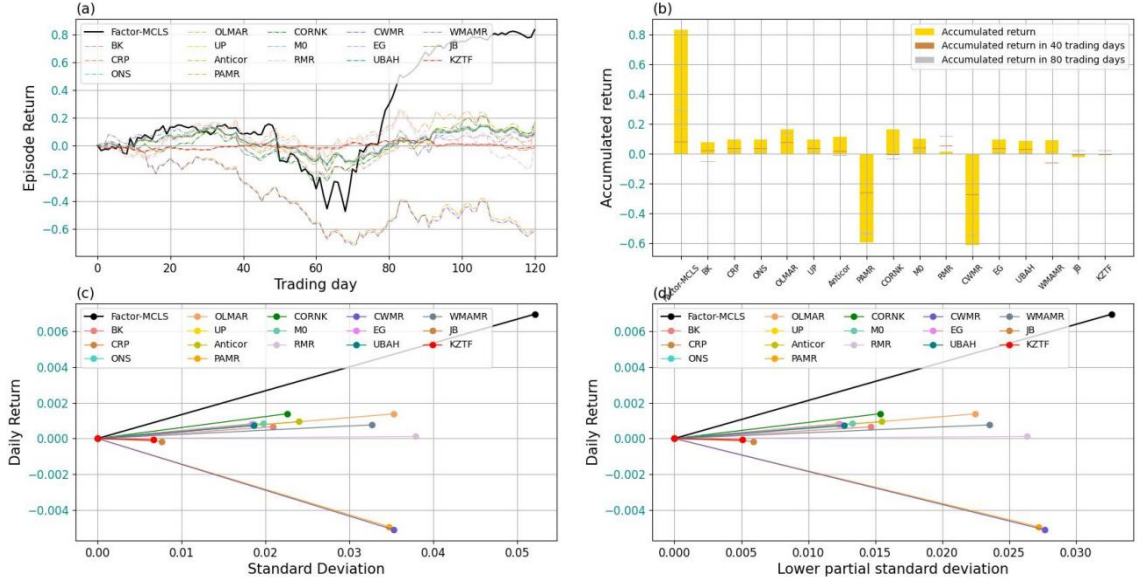

**Fig 8:** Performance Assessment of Portfolio Selection Algorithms in Experiment 2. The strategies and metrics depicted in this figure align with those presented in Figure 6.

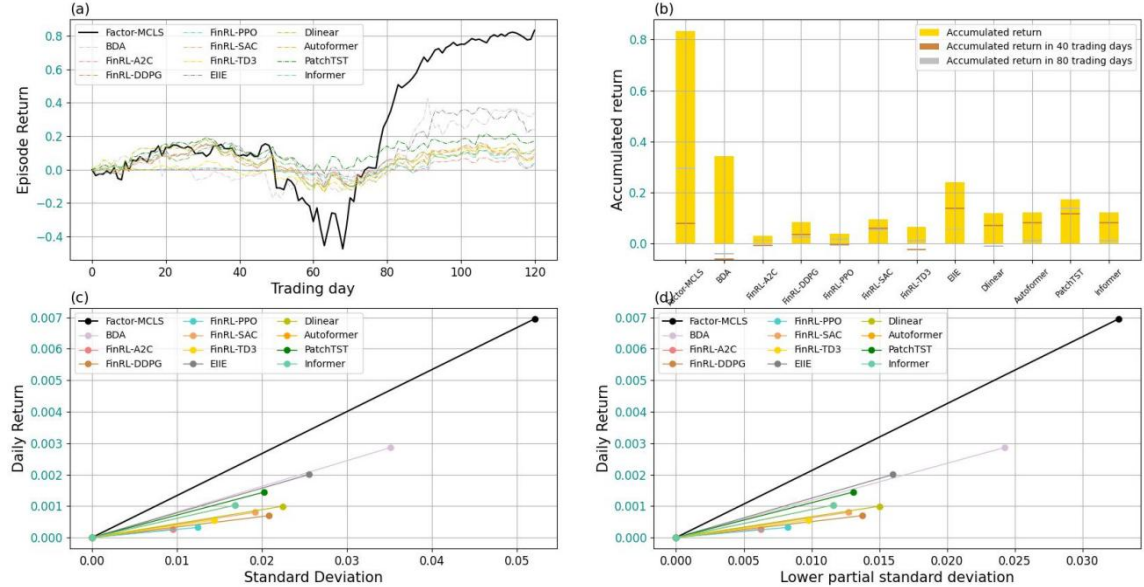

**Fig 9:** Performance Assessment of Portfolio Selection Strategies in Experiment 2. The strategies and metrics in this figure align with those in Figure 7.

## 5.7 Ablation study

To illustrate the value of employing a multi-critic framework and the risk constraint term $\Pi(\theta)$ in the training process, we conduct the following ablation studies. Specifically, we conduct two distinct ablation studies as described in Section 5.1. In each ablation study, we perform trajectory analysis on significant indices proposed in Section 5.2 during the training process and conduct comparative experiments in the back-testing environments.

**The necessity of the risk constrain term in the training objective function**

To demonstrate the necessity of incorporating the risk constraint term $\Pi(\theta)$ in the training objective function of the policy network, we develop a variant of Factor-MCLS specifically in this ablation experiment. In this variant, we ablate the risk constraint term in the training objective function. This variant is referred to as Learning System Variant 1 (LSV1). During the training process of LSV1, the trajectories of the significant indices defined in Section 5.2 are presented in Figures 10 and 11. The performance of LSV1, as measured by the indices outlined in Section 5.5 from the back-testing experiment, is summarized in Table 3. The numerical change trajectories of key indices shown in Fig. 10 reveal that the value of accumulated variance $AV^{(tra)}$ (depicted in subplot (c)) is over twice the converged value observed when using the risk constraint term $\Pi(\theta)$ (shown in Fig. 4, subplot (c)). This suggests that, despite the substantial improvement in profitability achieved Factor-MCLS during training after removing the risk constraint term $\Pi(\theta)$, there is no significant enhancement in the system's risk control capability. Simultaneously, the convergence speed of the loss function $\mathcal{L}_{Q_{(Va)}}(\omega_{(Va)})$ and $\mathcal{L}_{Q_{(Co)}}(\omega_{(Co)})$ used to train the critic networks $Q_{\omega_{(Va)}}(s_i, a_i)$ and $Q_{\omega_{(Co)}}(s_i, a_i)$ within the multi-critic framework (as illustrated in Fig. 11, subplots (d) and (e)) is significantly lower as compared to the convergence speed of the loss function $\mathcal{L}_{Q_{(Va)}}(\omega_{(Va)})$ and $\mathcal{L}_{Q_{(Co)}}(\omega_{(Co)})$ (as demonstrated in Fig. 5, subplots (d) and (e)) when the risk constraint term $\Pi(\theta)$ is employed. Consequently, we may conclude that incorporating the risk constraint term $\Pi(\theta)$ into the training objective function significantly enhances the risk control ability of Factor-MCLS. Efficient management of the risks encountered by Factor-MCLS during dynamic portfolio optimization also ensures the convergence of the training objective function when training the critic networks. This suggests that the risk constraint term $\Pi(\theta)$ can alleviate the training difficulty of the critic networks within the multi-critic framework.

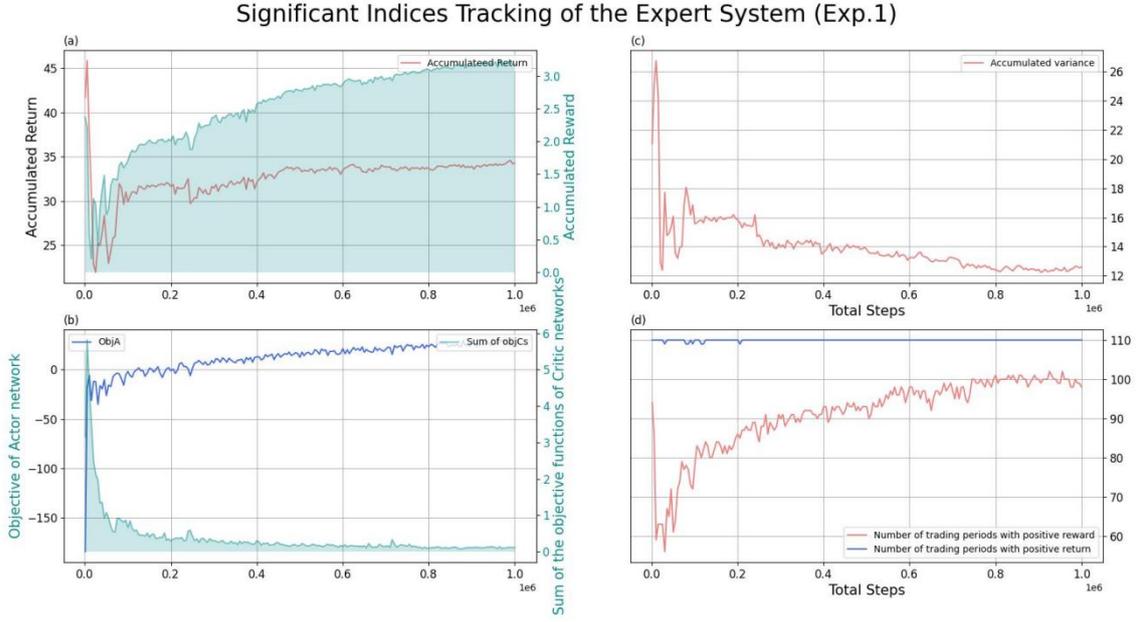

**Fig 10:** Significant indices tracked in the training process of LSV1 in experiment 1. The indices being tracked, as illustrated in the figure, are aligned with those presented in Figure 4.

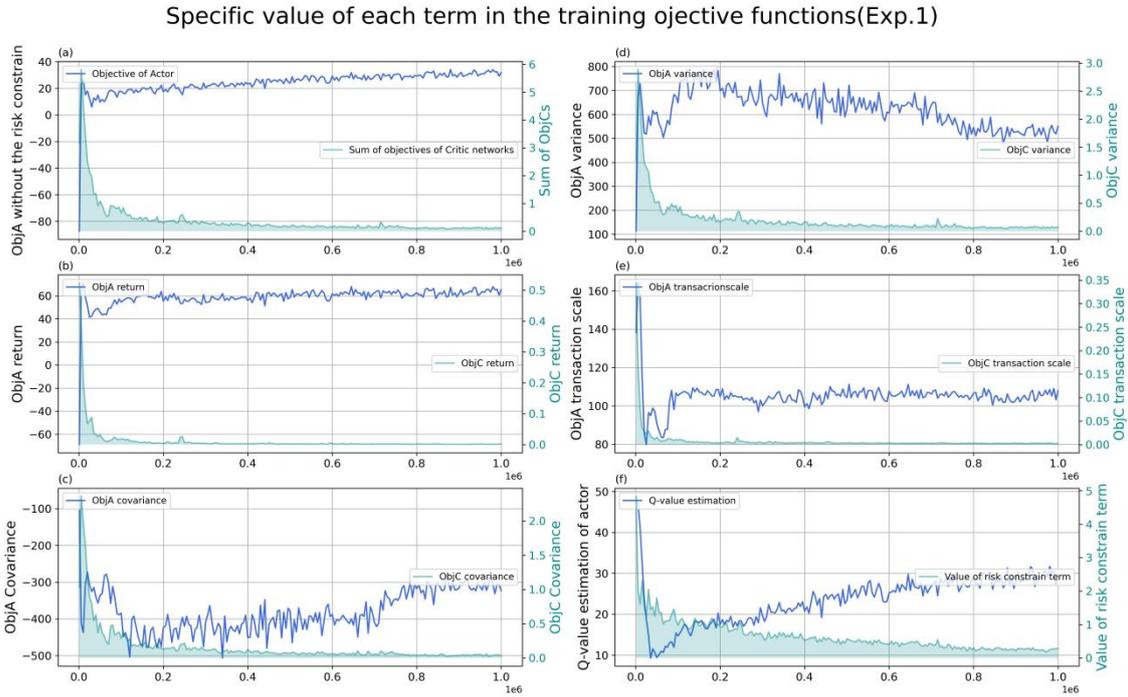

**Fig 11:** The specific values of the terms in the training objective functions. The indices being tracked, as illustrated in the figure, are aligned with those presented in Figure 5.

| Exp. | Name | Performance Metrics | | | | | |
|---|---|---|---|---|---|---|---|
| | | AR | DR | Std | SR | LStd | STR |
| 1 | **Factor-MCLS** | **0.46628** | **0.00389** | **0.02856** | **0.13604** | 0.02601 | **0.22273** |
| | LSV1 | 0.03555 | 0.00030 | 0.04430 | 0.00669 | 0.04227 | 0.00960 |
| 2 | **Factor-MCLS** | 0.83419 | 0.00695 | 0.05209 | 0.13347 | 0.04739 | 0.21282 |

| | | | | | | |
|---|---|---|---|---|---|---|
| | LSV1 | -0.16471 | -0.00137 | 0.10298 | -0.01333 | 0.10905 | -0.01810 |

**Table 4:** The experimental results of ablation study 1. The best results for each experiment are highlighted in bold.

The back-testing results presented in Table 3 demonstrate that Factor-MCLS achieves significantly higher accumulated returns (AR), daily returns (DR), Sharpe ratios (SR), and Sortino ratios (STR) across various experiments compared to LSV1. This suggests that the removal of the risk constraint term $\Pi(\theta)$ compromises the risk control ability of Factor-MCLS, which in turn impacts the out-of-sample generalization of the system's policy regarding profitability and risk-adjusted returns.

**The necessity of applying the critic networks in the muti-critics framework to construct the training objective function**

In this ablation experiment, the function defined in Equation (19) is used as a substitute for the original training objective function of the DRL agent's policy network, as specified in Equation (13). This variant is referred to as Learning System Variant 2 (LSV2). Here, we demonstrate that, in terms of ensuring the DRL agent's control over portfolio risk while maintaining the generalization capability of the dynamic portfolio optimization policy, the objective function we develop for training the policy network outperforms the objective function used in traditional actor-critic algorithms.

During the training process of LSV2, the trajectories of the significant indices defined in Section 5.2 are presented in Figure 12. The performance of LSV2, as measured by the indices outlined in Section 5.5 from the back-testing experiment, is summarized in Table 5. The trajectory of key index values presented in Fig. 12 indicates that the accumulated variance $AV^{(tra)}$ associated with LSV2 (as shown in Fig. 12, subplot (c)) exceeds twice the converged value achieved by Factor-MCLS (shown in Fig. 4, subplot (c)), which is trained using the objective function $\mathcal{L}_\pi(\theta)$ defined in Equation (13). This indicates that, although the profitability of LSV2 is significantly enhanced during training with the traditional actor-critic algorithm, its risk control abilities could not be effectively improved. In summary, while employing the traditional actor-critic algorithm, the inclusion of a risk calibration term in the reward function cannot significantly enhance the risk control capabilities of Factor-MCLS. Throughout the training process, the agent continues to prioritize improving profitability, resulting in a neglect of risk control enhancement. This indicates that when the reward factor vector is unavailable, Factor-MCLS fails to adequately account for all the influencing factors within the reward function. As a result, it cannot effectively balance the improvement of both profitability and risk control.

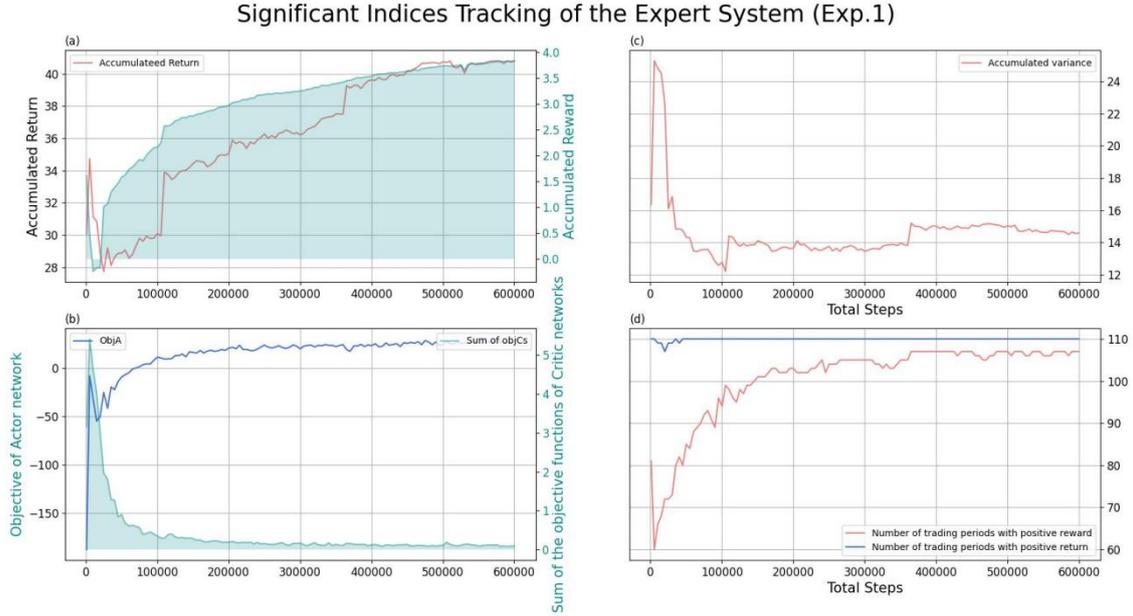

**Fig 12:** Significant indices tracked in the training process of LSV2 in experiment 1. The indices being tracked, as illustrated in the figure, are aligned with those presented in Figure 4.

| Exp. | Name | Performance Metrics | | | | | |
|---|---|---|---|---|---|---|---|
| | | AR | DR | Std | SR | LStd | STR |
| 1 | **Factor-MCLS** | **0.46628** | **0.00389** | 0.02856 | **0.13604** | 0.02601 | **0.22273** |
| | LSV2 | -0.33151 | -0.00276 | 0.09625 | -0.02870 | 0.08910 | -0.04385 |
| 2 | **Factor-MCLS** | **0.83419** | **0.00695** | 0.05209 | **0.13347** | 0.04739 | **0.21282** |
| | LSV2 | -2.14804 | -0.01790 | 0.26251 | -0.06819 | 0.28057 | -0.08876 |

**Table 5:** The experimental results of ablation study 2. The best results for each experiment are highlighted in bold.

The experimental results presented in Table 5 indicate that Factor-MCLS achieves significantly higher accumulated returns (AR), daily returns (DR), Sharpe ratios (SR), and Sortino ratios (STR) across various back-testing experiments compared to LSV2. This indicates that training the policy network of the Deep Reinforcement Learning (DRL) agent using the traditional Deep Deterministic Policy Gradient (DDPG) algorithm's objective function may result in insufficient enhancement of the learning system's risk control capabilities. Such inadequacies can adversely affect the out-of-sample generalization of the policy concerning profitability and return per unit of risk.

# 6. Conclusion and future work

This paper proposes a method for achieving comprehensive learning of the factors affecting portfolio return and risk within the training environment when applying reinforcement learning algorithms in dynamic portfolio optimization. Specifically, we propose a novel reward factor matrix that provides a comprehensive analysis of all the factors influencing portfolio risk and return. Furthermore, in our research, we propose a novel DRL-based learning system. In this learning system, we develop a multi-critic framework that facilitates comprehensive learning of the various influencing factors within the reward factor matrix. Furthermore, we have established a risk constraint term derived from the critic networks within the multi-critic framework. This risk constraint term allows investors to intervene in the training of the DRL agent in the learning system according to their risk aversion for the assets in the portfolio. During the training process, our Factor-MCLS exhibits exceptional risk control abilities. Concurrently, our training algorithm significantly enhances both profitability and risk-adjusted profitability while maintaining outstanding performance in risk control. In the back-testing environment, it significantly outperforms the comparative strategies based on traditional capital growth theory and those using machine learning algorithms. This suggests that Factor-MCLS can maintain exceptional generalization ability in terms of profitability and risk-adjusted profitability in out-of-sample settings. The results of the ablation study indicate that the training objective function and risk constraint term, developed based on the critic network within the multi-critic framework, play an essential role in ensuring the risk control abilities of the learning system in the training environment and its generalization ability in out-of-sample settings.

Despite the positive empirical outcomes achieved, Factor-MCLS still has two major limitations. First, while actively intervening in the training of the DRL agent within it using risk constraint terms, we lack a model that can determine the risk aversion of each asset in the portfolio based on the investor's evaluation of arbitrage opportunities among various assets. Second, the BDA, functioning as an auxiliary agent, has a policy function constructed based on the assumption of a normal distribution. However, in real markets, the return distribution typically displays heavier tails and sporadic high peaks, which cannot be accurately represented by a Gaussian distribution.

In future work, in order to overcome the shortcomings of the current model, we will make the following research. In the policy function of the auxiliary agent, we will construct a Bayesian model based on distributions that can describe heavier tails and high peaks (i.e., Elliptical distributions). Furthermore, to further improve the out-of-sample performance in profitability, we will investigate developing a suitable model using machine learning algorithms to determine the optimal risk aversion of each portfolio asset based on the arbitrage opportunities of those assets. By assigning appropriate risk aversion levels to the various assets within the portfolio, we may further refine the training of the DRL agent in Factor-MCLS, thereby enhancing the generalization ability of its policy.

**Data availability:** This research uses only publicly available American stock data from the Yahoo Finance platform. All datasets generated during and/or analyzed during the current study, models, or codes that support the findings of this study are available from the corresponding author upon reasonable request.

**Declarations Conflict of interest:** The authors declare that they have no known competing financial interests or personal relationships that could have appeared to influence the work reported in this paper.